\documentclass{article}

\usepackage{microtype}
\usepackage{graphicx}
\usepackage{subcaption} 
\usepackage{booktabs} 
\usepackage{algorithm}
\usepackage{algorithmic}
\usepackage{amssymb}
\usepackage{epstopdf}
\usepackage{multirow}
\newtheorem{assumption}{Assumption}
\newtheorem{lemma}{Lemma}

\newtheorem{theorem}{Theorem}
\newtheorem{corollary}{Corollary}
\newtheorem{proof}{Proof}
\usepackage{amsmath}
\usepackage{mathtools}
\renewenvironment{proof}{{ \par \noindent \textit{Proof:}}}{}

\usepackage{hyperref}

\usepackage[accepted]{icml2018}

\icmltitlerunning{Decoupled Parallel Backpropagation with Convergence Guarantee}

\begin{document}

\twocolumn[
\icmltitle{Decoupled Parallel Backpropagation with Convergence Guarantee
}



\icmlsetsymbol{equal}{*}

\begin{icmlauthorlist}
\icmlauthor{Zhouyuan Huo}{pitt}
\icmlauthor{Bin Gu}{pitt}
\icmlauthor{Qian Yang}{pitt}
\icmlauthor{Heng Huang}{pitt}
\end{icmlauthorlist}

\icmlaffiliation{pitt}{Department of Electrical and Computer Engineering, University of Pittsburgh, Pittsburgh, United States}

\icmlcorrespondingauthor{Heng Huang}{heng.huang@pitt.edu}

\icmlkeywords{Machine Learning, ICML}

\vskip 0.3in
]



\printAffiliationsAndNotice{}  

\begin{abstract}
Backpropagation algorithm is indispensable for the training of feedforward neural networks. 
It requires propagating error gradients sequentially from the output layer all the way back to the input layer.
The backward locking in backpropagation algorithm constrains us from updating network layers in parallel and fully leveraging the computing resources. Recently, several algorithms have been proposed for breaking the backward locking. However, their performances degrade seriously when networks are deep.  In this paper,  we propose decoupled parallel backpropagation algorithm for deep learning optimization with convergence guarantee.  Firstly, we decouple the backpropagation algorithm using delayed gradients, and show that the backward locking is removed when we split the networks into multiple modules.
Then, we utilize decoupled parallel backpropagation in two stochastic methods and prove that our method guarantees convergence to critical points for the non-convex problem.
 Finally, we perform experiments for training deep convolutional neural networks on benchmark datasets. The experimental results not only confirm our theoretical analysis, but also demonstrate that the proposed method can achieve significant speedup without loss of accuracy. Code is available at  \href{https://github.com/slowbull/DDG}{https://github.com/slowbull/DDG}.
\end{abstract}

\section{Introduction}
\begin{figure}[t]
	\centering
	\includegraphics[width=3.in]{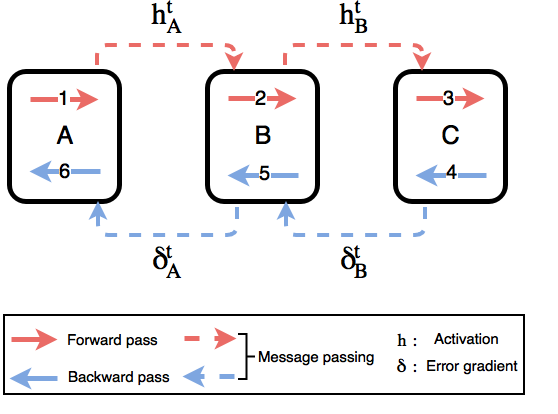}
	\caption{We split a multilayer feedforward neural network into three modules. Each module is a stack of layers. Backpropagation algorithm requires running forward pass (from $1$ to $3$) and backward pass (from $4$ to $6$) in sequential order. For example, module $A$ cannot perform step $6$ before receiving $\delta_A^t$ which is an output of step $5$ in module B. }
	\label{intro}
\end{figure}

We have witnessed a series of breakthroughs in computer vision using deep convolutional neural networks \cite{lecun2015deep}. 
Most neural networks are trained using stochastic gradient descent (SGD) or its variants in which the gradients of the networks are computed by backpropagation algorithm \cite{rumelhart1988learning}. As shown in Figure \ref{intro}, the backpropagation algorithm consists of two processes, the forward pass to compute prediction and the backward pass to compute gradient and update the model.
After computing prediction in the forward pass, backpropagation algorithm requires propagating error gradients from the top (output layer) all the way back to the bottom (input layer). Therefore, in the backward pass, all layers, or more generally, modules, of the network are locked until their dependencies have executed. 

The backward locking  constrains us from updating models in parallel and fully leveraging the computing resources. It has been shown in practice \cite{krizhevsky2012imagenet,simonyan2014very,szegedy2015going,he2016deep,huang2016densely} and in theory \cite{eldan2016power,telgarsky2016benefits,bengio2009learning} 
that depth is one of the most critical factors contributing to the success of deep learning. From AlexNet with $8$ layers \cite{krizhevsky2012imagenet} to ResNet-101 with more than one hundred layers \cite{he2016deep}, the forward and backward time grow from ($4.31$ms and $9.58$ms) to  ($53.38$ms and $103.06$ms) when we train the networks on Titan X with the input size of $16 \times 3 \times 224 \times 224$ \cite{benchmark}. Therefore, parallelizing the backward pass can greatly reduce the training time when the backward time is  about twice of the forward time. We can easily split a deep neural network into  modules like Figure \ref{intro} and distribute them across multiple GPUs. However,  because of the backward locking, all GPUs are idle before receiving error gradients from dependent modules in the backward pass.

 There have been several algorithms proposed for breaking the backward locking. For example,  \cite{jaderberg2016decoupled,czarnecki2017understanding} proposed to remove the lockings in backpropagation by employing additional neural networks to approximate error gradients. In the backward pass, all modules use the  synthetic gradients to update weights of the model without incurring any delay.
\cite{nokland2016direct,balduzzi2015kickback} broke the local dependencies between successive layers and  made all hidden layers receive error information from the output layer directly.   
In \cite{carreira2014distributed,taylor2016training}, the authors loosened the exact connections between layers by introducing auxiliary variables. In each layer, they imposed equality constraint between the auxiliary variable and activation, and optimized the new problem using Alternating Direction Method which is easy to parallel.
However, for the convolutional neural network, the performances of all above methods are much worse than backpropagation algorithm  when the network is deep.

In this paper, we focus on breaking the backward locking in  backpropagtion algorithm for training feedforward neural networks, such that we can update  models in parallel 
without loss of accuracy. The main contributions of our work are as follows: 
\begin{itemize}
	\item Firstly, we decouple the backpropagation using delayed gradients in Section \ref{sec_alg} such that all modules of the network can be updated in parallel without  backward locking.
	\item Then,  we propose two stochastic algorithms using decoupled parallel backpropagation in Section \ref{sec_alg} for deep learning optimization.  
	\item  We also provide convergence analysis for the proposed method in Section \ref{sec_conv} and prove that it guarantees convergence to critical points for the non-convex problem.
	\item 	 Finally, we perform experiments for training deep convolutional neural networks in Section \ref{sec_exp}, experimental results verifying that the proposed method can significantly speed up the training  without loss of accuracy.
\end{itemize}

 \section{Backgrounds}
We begin with a brief overview of the backpropagation algorithm for the optimization of neural networks. Suppose that we want to train a feedforward neural network with $L$ layers, each layer taking an input $h_{l-1}$ and producing an activation $h_{l} = F_l(h_{l-1}; w_l)$ with weight $w_l$. Letting $d$ be the dimension of weights in the network, we have $w=[w_1,w_2,...,w_L] \in \mathbb{R}^d$.  Thus, the output of the network can be represented as $h_L = F(h_0; w)$, where $h_0$ denotes the input data $x$. Taking a loss function $f$ and targets $y$, the training problem is as follows:
\begin{eqnarray}
\min\limits_{w=[w_1,...,w_L]} f(F(x; w),y).
\label{obj}
\end{eqnarray}
In the following context, we use $f(w)$ for simplicity. 

Gradients based methods are widely used for deep learning optimization \cite{robbins1951stochastic,qian1999momentum,hinton2012neural,kingma2014adam}.  In iteration $t$, we put a data sample $x_{i(t)}$ into the network, where $i(t)$ denotes the index of the sample. According to stochastic gradient descent (SGD), we update the weights of the network through:
 \begin{eqnarray}
 	w_l^{t+1} = w_l^{t} - \gamma_t  \left[\nabla f_{l, x_{i(t)}} (w^t)  \right]_l, \hspace{0.1cm} \forall l \in \{1,2,...,L\}
 	\label{sgd}
 \end{eqnarray}
 where $\gamma_t$ is the stepsize and $\nabla f_{l, x_{i(t)}} (w^t)\in \mathbb{R}^d$ is the gradient of the loss function  (\ref{obj})  with respect to the weights at layer $l$ and data sample $x_{i(t)}$, all the coordinates in other than layer $l$ are $0$. 
 We always utilize backpropagation algorithm to compute the gradients \cite{rumelhart1988learning}. The backpropagation algorithm consists of two passes of the network:
in the forward pass, the activations of all layers are calculated from $l=1$ to $L$  as follows:
\begin{eqnarray}
	h_l^t = F_l(h^t_{l-1}; w_l)
	\label{forward};
\end{eqnarray}
in the backward pass, we apply chain rule for gradients and repeatedly propagate error gradients through the network from the output layer $l=L$ to the input layer $l=1$:
\begin{eqnarray}
\frac{\partial f(w^t)}{\partial w^t_{l}}& =& \frac{\partial h^t_l}{\partial w^t_l}  \frac{\partial f(w^t)  }{\partial h_l^t},\label{backward1} \\
\frac{\partial f(w^t)}{\partial h^t_{l-1}} &=& \frac{\partial h^t_l}{\partial h^t_{l-1}}  \frac{\partial f(w^t)  }{\partial h^t_l},
	\label{backward2}
\end{eqnarray}
where we let  $\nabla f_{l, x_{i(t)}} (w^t) = \frac{\partial f(w^t)}{\partial w^t_{l}}$.
From equations (\ref{backward1}) and (\ref{backward2}), it is obvious that the computation in layer $l$ is dependent on the error gradient $\frac{\partial f(w^t)  }{\partial h_l^t}$ from layer $l+1$. Therefore, the backward locking constrains all layers from updating before receiving error gradients from the dependent layers. When the network is very deep or distributed across multiple resources, the backward locking is the main bottleneck in the training process. 
 
 \begin{figure*}[th]
 	\centering
 	\includegraphics[width=4.5in]{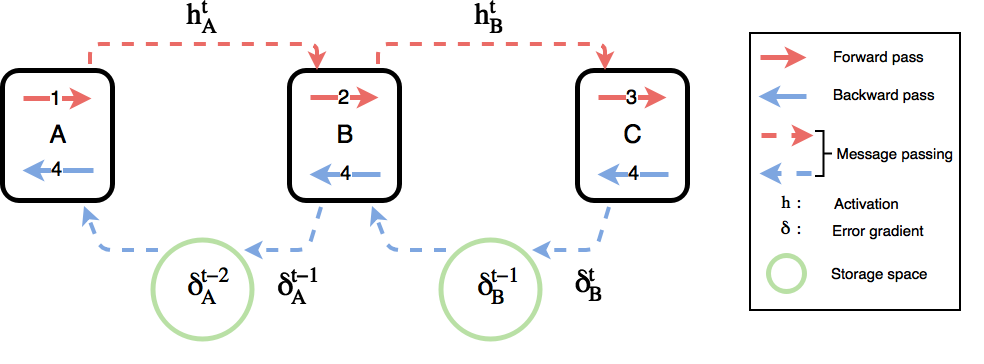}
 	\caption{We split a multilayer feedforward neural network into three modules (A, B and C), where each module is a stack of layers. After executing the forward pass (from $1$ to $3$) to predict, our proposed method allows all modules to run backward pass ($4$) using delayed gradients without locking. Particularly, module $A$ can perform the backward pass using the stale error gradient $\delta_A^{t-2}$. Meanwhile, It also receives $\delta_A^{t-1}$ from module $B$ for the update of the next iteration. }
 	\label{algo}
 \end{figure*}

\section{Decoupled Parallel Backpropagation}
\label{sec_alg}
In this section, we propose to decouple the backpropagation algorithm using delayed gradients (DDG). Suppose we split a $L$-layer feedforward neural network to $K$ modules, such that the weights of the network are divided into $K$ groups. Therefore, we have ${w}= [w_{\mathcal{G}(1)},w_{\mathcal{G}(2)},..., w_{\mathcal{G}(K)}]$ where $\mathcal{G}(k)$ denotes layer indices in the group $k$.

\subsection{Backpropagation Using Delayed Gradients}
In iteration $t$, data sample $x_{i(t)}$ is input to the network.  We run the forward pass from module $k=1$ to $k=K$. In each module, we compute the activations in sequential order as equation (\ref{forward}). In the backward pass, all modules except the last one have delayed error gradients in store such that they can execute the backward computation without locking. The last module updates with the up-to-date gradients. In particular, module $k$ keeps the stale error gradient $\frac{\partial f(w^{t-K+k})}{\partial h_{L_k}^{t-K+k}}$, where $L_k$ denotes the last layer in module $k$.  Therefore, the backward computation in module $k$ is as follows:
\begin{eqnarray}
\frac{\partial f(w^{t-K+k})}{\partial w^{t-K+k}_{l}}& =& \frac{\partial h^{t-K+k}_l}{\partial w^{t-K+k}_l}  \frac{\partial f(w^{t-K+k}) }{\partial h_l^{t-K+k}},\label{backward3} \\
\frac{\partial f(w^{t-K+k})}{\partial h^{t-K+k}_{l-1}} &=& \frac{\partial h^{t-K+k}_l}{\partial h^{t-K+k}_{l-1}}  \frac{\partial f(w^{t-K+k})}{\partial h^{t-K+k}_l}.
\label{backward4}
\end{eqnarray}
where $\ell \in \mathcal{G}(k)$.  
Meanwhile, each module also receives error gradient from the dependent module for further computation. From (\ref{backward3}) and  (\ref{backward4}), we can know that the stale error gradients in all modules are of different time delay.
From module $k=1$ to $k=K$, their corresponding time delays are from $K-1$ to $0$. Delay $0$ indicates that the gradients are up-to-date.  In this way, we break the backward locking and achieve parallel update in the backward pass. Figure \ref{algo} shows an example of the decoupled backpropagation, where error gradients $ \delta \coloneqq \frac{\partial f(w)}{\partial h}$. 

\subsection{Speedup of Decoupled Parallel Backpropagation}
When $K=1$, there is no time delay and the proposed method is equivalent to the backpropagation algorithm. 
 When $K\neq 1$, we can distribute the network across multiple GPUs and fully leverage the computing resources. Table \ref{time} lists the computation time when we sequentially allocate the network across $K$ GPUs. When $\mathcal{T}_F$ is necessary to compute accurate predictions, we can accelerate the training by reducing the backward time. 
 Because $\mathcal{T}_B$ is much large than $\mathcal{T}_F$, we can achieve huge speedup even $K$ is small.

\begin{table}[t]
	\caption{Comparisons of computation time when the network is sequentially distributed across $K$ GPUs. 
		$\mathcal{T}_F$ and $\mathcal{T}_B$ denote the forward and backward time for backpropagation algorithm. }
		\center
	\begin{tabular}{c|c}
		\hline
		{Method} & {Computation Time} \\\hline
		Backpropagation         & $\mathcal{T}_F+\mathcal{T}_B$  \\\hline 
		DDG        & $\mathcal{T}_F+\frac{\mathcal{T}_B}{K}$ \\\hline 
	\end{tabular}
	\label{time}
\end{table}
\textbf{Relation to model parallelism:} Model parallelism usually refers to filter-wise parallelism \cite{yadan2013multi}. For example, we split a convolutional layer with $N$ filters into two GPUs, each part containing $\frac{N}{2}$ filters. Although the filter-wise parallelism accelerates the training when we distribute the workloads across multiple GPUs, it still suffers from the backward locking.  We can think of DDG algorithm as layer-wise parallelism. It is also easy to combine filter-wise parallelism with layer-wise parallelism for further speedup.

\subsection{Stochastic Methods Using  Delayed Gradients}
After computing the gradients of the loss function with respect to the weights of the model, we update the model using delayed gradients. Letting $\nabla f_{{\mathcal{G}(k)}, x_{i(t-K+k)}} \left(w^{t-K+k}\right):=$
\begin{eqnarray}
\left\{\begin{matrix}
\sum\limits_{ l \in \mathcal{G}(k)} \frac{\partial f(w^{t-K+k})}{\partial w^{t-K+k}_{l}}  & \text{if  } t-K+k \geq 0\\ 
0& \text{otherwise} 
\end{matrix}\right.,
\end{eqnarray}
for any $ k \in \{1,2,...,K\} $,  we update the weights in module $k$ following SGD:
\begin{eqnarray} 
w^{t+1}_{\mathcal{G}(k)} = w^t_{\mathcal{G}(k)} - \gamma_t [\nabla f_{{\mathcal{G}(k)}, x_{i(t-K+k)}} \left(w^{t-K+k}\right)]_{\mathcal{G}(k)}.
\label{ddg}
\end{eqnarray}
where $\gamma_t$ denotes stepsize. 
Different from SGD, we update the weights with delayed gradients. Besides, the delayed iteration $(t-K+k)$ for group  $k$ is also deterministic.  We summarize the proposed method in 
Algorithm \ref{alg}.

\begin{algorithm}[t]
	\caption{SGD-DDG}
	\begin{algorithmic}[1]
		\REQUIRE~~\\
		 Initial weights $w^0 = [w^0_{\mathcal{G}(1)}, ..., w^0_{\mathcal{G}(K)}] \in \mathbb{R}^d$;\\
		 Stepsize sequence $\{\gamma_t\}$;
		\FOR{$t=0,1,2,\dots, T-1$}
		\FOR{$k=1, \dots, K$ \textbf{in parallel}}
		\STATE Compute delayed gradient: \\
		$\hspace{1cm}g_k^t \leftarrow \left[\nabla f_{{\mathcal{G}(k)}, x_{i(t-K+k)}} \left(w^{t-K+k}\right)\right]_{\mathcal{G}(k)};$
		\STATE Update weights:\\
		$\hspace{1cm}w^{t+1}_{\mathcal{G}(k)} \leftarrow  w^t_{\mathcal{G}(k)} - \gamma_t \cdot g_k^t;$
		\ENDFOR
		\ENDFOR
	\end{algorithmic}
	\label{alg}
\end{algorithm}

\begin{algorithm}[t]
	\caption{Adam-DDG}
	\begin{algorithmic}[1]
		\REQUIRE~~\\
		Initial weights: $w^0 = [w^0_{\mathcal{G}(1)}, ..., w^0_{\mathcal{G}(K)}] \in \mathbb{R}^d$;\\
		Stepsize: $\gamma$;  Constant $\epsilon = 10^{-8}$;\\
		Exponential decay rates: $\beta_1=0.9$ and $\beta_2 = 0.999$ ;\\
		First moment vector: $m^0_{\mathcal{G}(k)}\leftarrow 0, \forall k \in \{1,2,...,K\}$; \\
		Second moment vector: $v_{\mathcal{G}(k)}^0 \leftarrow 0, \forall k \in \{1,2,...,K\}$; \\
		\FOR{$t=0,1,2,\dots, T-1$}
		\FOR{$k=1, \dots, K$ \textbf{in parallel}}
		\STATE Compute delayed gradient: \\
		$\hspace{1cm}g_k^t \leftarrow \left[\nabla f_{{\mathcal{G}(k)}, x_{i(t-K+k)}} \left(w^{t-K+k}\right)\right]_{\mathcal{G}(k)};$
		\STATE Update biased first moment estimate: \\
		 		$\hspace{1cm}m^{t+1}_{\mathcal{G}(k)} \leftarrow \beta_1 \cdot m_{\mathcal{G}(k)}^t + (1-\beta_1) \cdot g_k^t$
		\STATE Update biased second moment estimate: \\
		$\hspace{1cm}v^{t+1}_{\mathcal{G}(k)} \leftarrow \beta_2 \cdot v_{\mathcal{G}(k)}^t + (1-\beta_2) \cdot (g_k^t)^2$
		\STATE   Compute bias-correct first moment estimate:\\
		 		$\hspace{1cm}\hat m^{t+1}_{\mathcal{G}(k)} \leftarrow  m^{t+1}_{\mathcal{G}(k)} / (1- \beta_1^{t+1})$
		\STATE   Compute bias-correct second moment estimate:\\
			$\hspace{1cm}\hat v^{t+1}_{\mathcal{G}(k)} \leftarrow  v^{t+1}_{\mathcal{G}(k)} / (1- \beta_2^{t+1})$ 
		\STATE Update weights: \\
		$\hspace{0.9cm} w^{t+1}_{\mathcal{G}(k)} \leftarrow w^t_{\mathcal{G}(k)} - \gamma \cdot \hat m_{\mathcal{G}(k)}^{t+1} / \left( \sqrt{\hat v_{\mathcal{G}(k)}^{t+1}}  + \epsilon\right)$
		\ENDFOR
		\ENDFOR
	\end{algorithmic}
	\label{alg2}
\end{algorithm}

Moreover, we can also apply the delayed gradients to other variants of SGD, for example Adam in Algorithm \ref{alg2}. In each iteration, we update the weights and moment vectors with delayed gradients.  We  analyze the convergence for Algorithm \ref{alg} in Section \ref{sec_conv}, which is the basis of analysis for other methods.

\section{Convergence Analysis}
\label{sec_conv}
In this section, we establish the convergence guarantees to critical points for Algorithm \ref{alg} when the problem is non-convex. Analysis shows that our method admits similar convergence rate to vanilla stochastic gradient descent \cite{bottou2016optimization}.
Throughout this paper, we make the following commonly used assumptions:
\begin{assumption}
\textbf{(Lipschitz-continuous gradient)}	The gradient of $f(w)$ is Lipschitz continuous with Lipschitz constant $L > 0$, such that $\forall  w, v \in \mathbb{R}^d$:
		\begin{eqnarray}
		\left\| \nabla f(w) - \nabla f(v) \right\|_2 \leq L \|w - v\|_2
 	\end{eqnarray}
		\label{lips}
\end{assumption}
\begin{assumption}
\textbf{(Bounded variance)} To bound the variance of the stochastic gradient, we assume the second moment of the stochastic gradient is upper bounded, such that there exists constant $M \geq 0$,  for any sample $x_i$ and $\forall w \in \mathbb{R}^d$: 
\begin{eqnarray}
\|\nabla f_{x_i} (w)\|^2_2&\leq & M
\end{eqnarray}
Because of the unnoised stochastic gradient $\mathbb{E} \left[\nabla f_{x_i} (w)\right] =\nabla f (w)  $ and the equation regarding variance $\mathbb{E} \left\|\nabla f_{x_i} (w) - \nabla f(w) \right\|^2_2 = \mathbb{E} \| \nabla f_{x_i} (w) \|^2_2 - \| \nabla f(w) \|^2_2,$  the variance of the stochastic gradient is guaranteed to be  less than  $M$.
\label{bg}
\end{assumption}

\begin{figure*}[th]
	\centering
	\begin{subfigure}[b]{0.24\textwidth}
		\centering
		\includegraphics[width=1.82in]{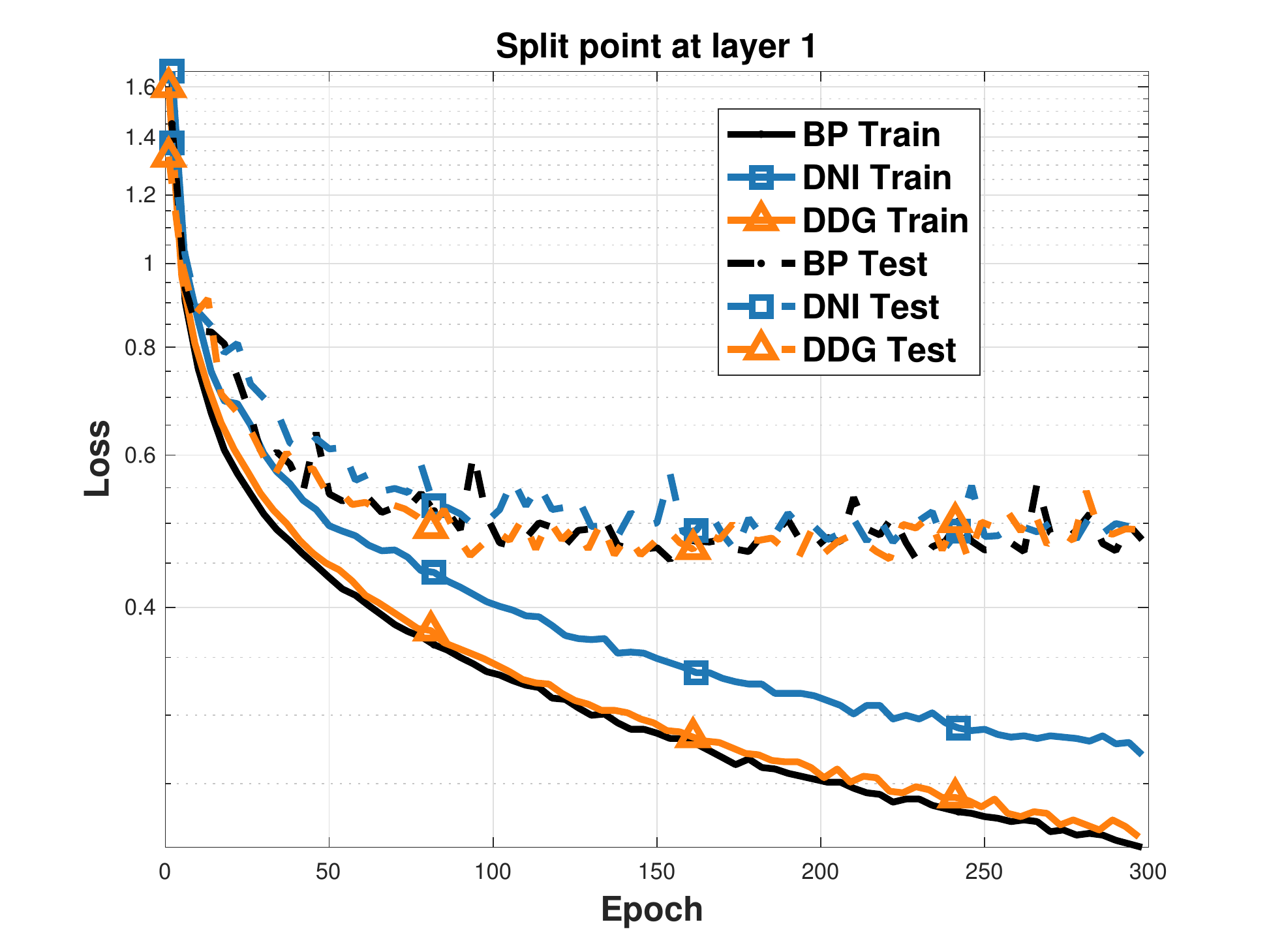}
	\end{subfigure}
	\begin{subfigure}[b]{0.24\textwidth}
		\centering
		\includegraphics[width=1.82in]{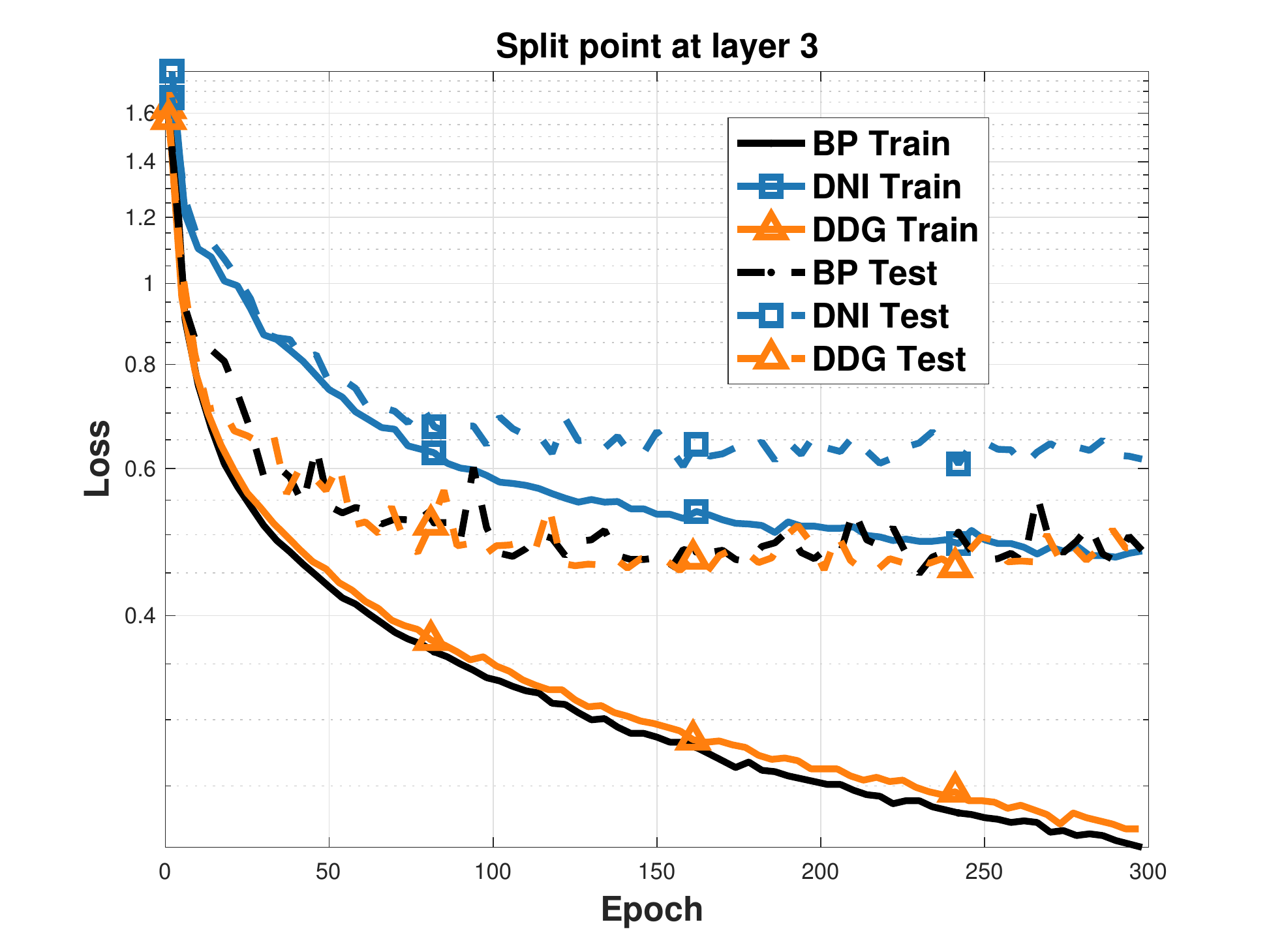}
	\end{subfigure}
	\begin{subfigure}[b]{0.24\textwidth}
		\centering
		\includegraphics[width=1.82in]{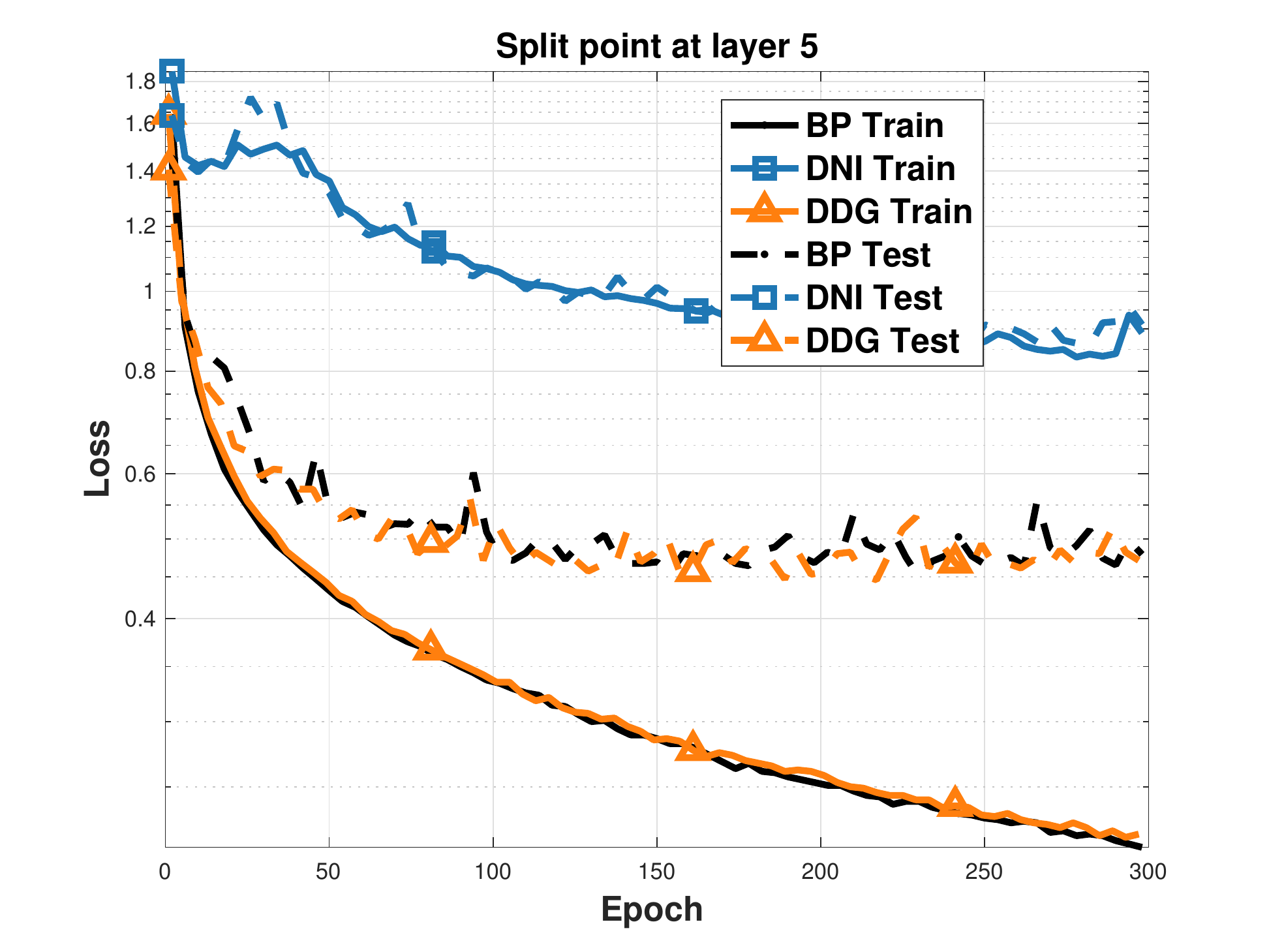}
	\end{subfigure}
	\begin{subfigure}[b]{0.24\textwidth}
		\centering
		\includegraphics[width=1.82in]{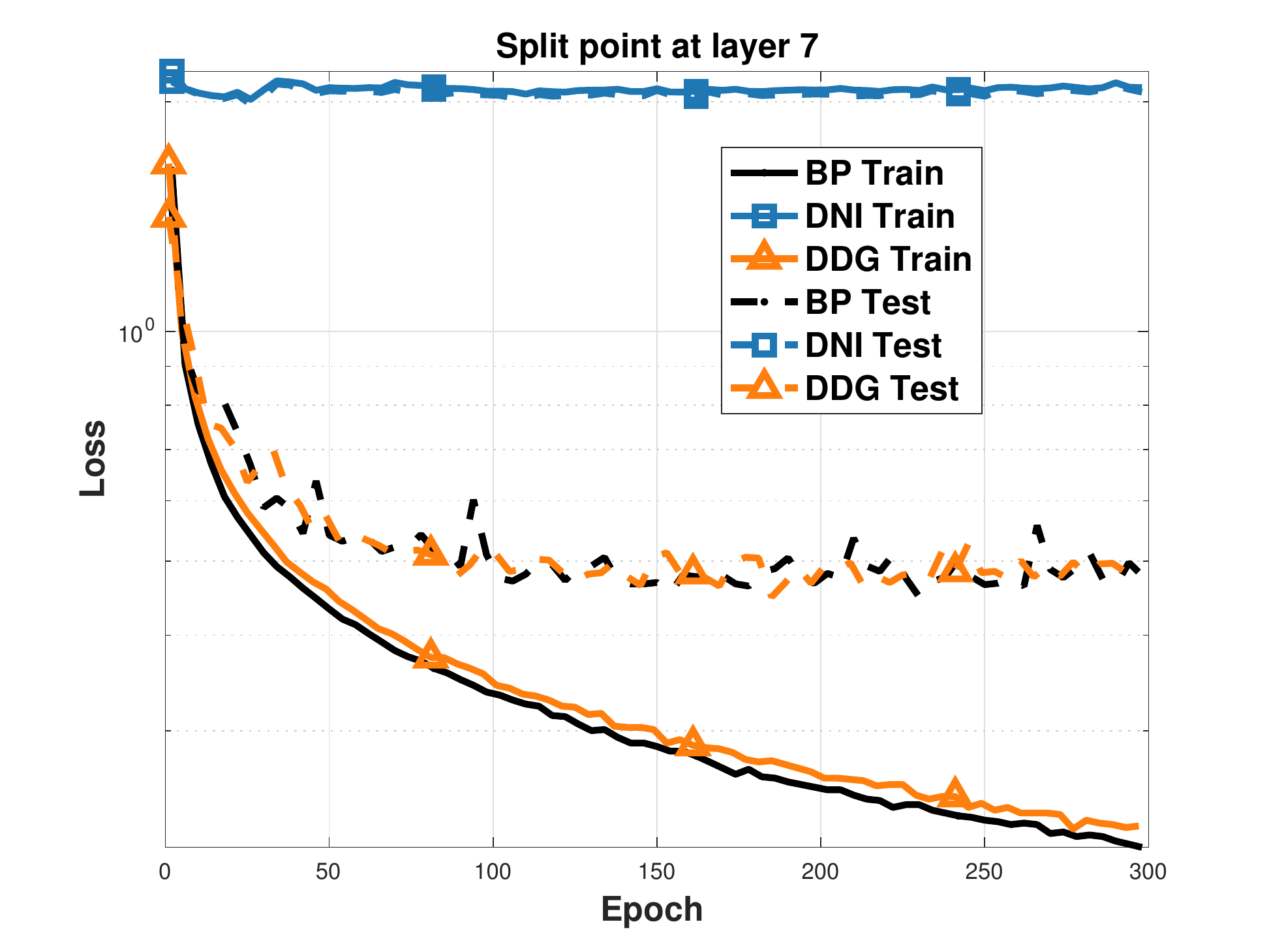}
	\end{subfigure}
	\begin{subfigure}[b]{0.24\textwidth}
		\centering
		\includegraphics[width=1.82in]{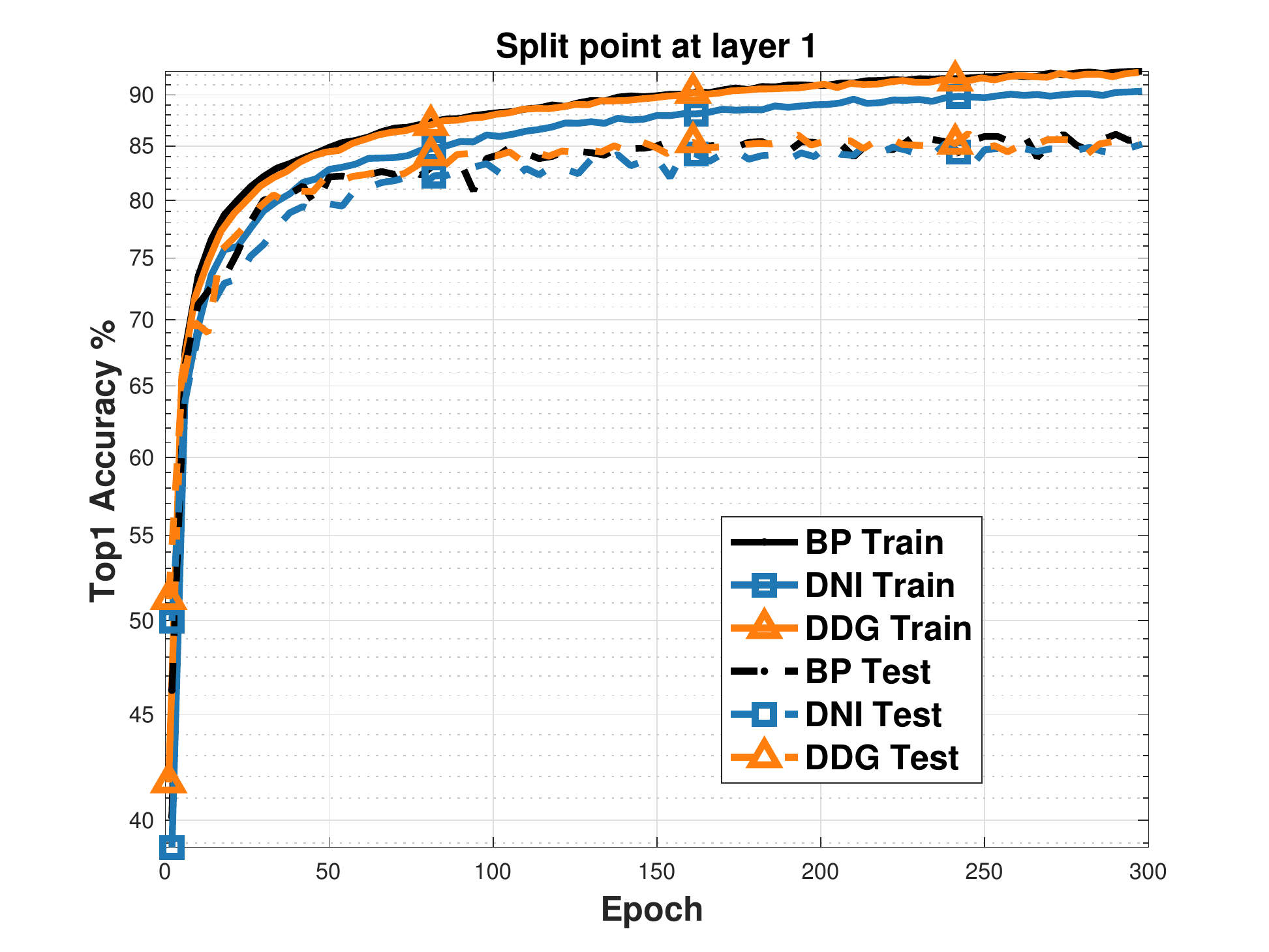}
	\end{subfigure}
	\begin{subfigure}[b]{0.24\textwidth}
		\centering
		\includegraphics[width=1.82in]{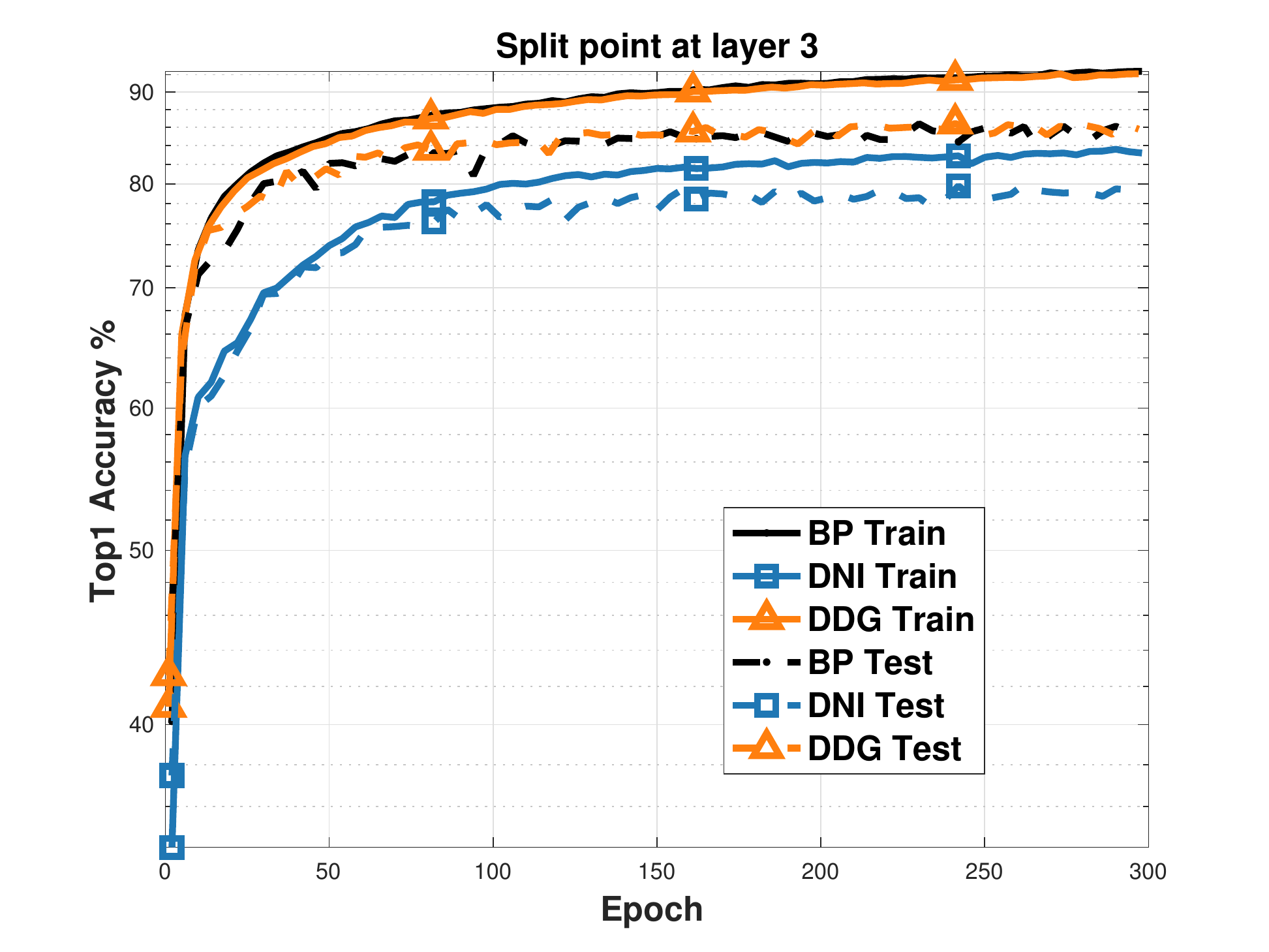}
	\end{subfigure}
	\begin{subfigure}[b]{0.24\textwidth}
		\centering
		\includegraphics[width=1.82in]{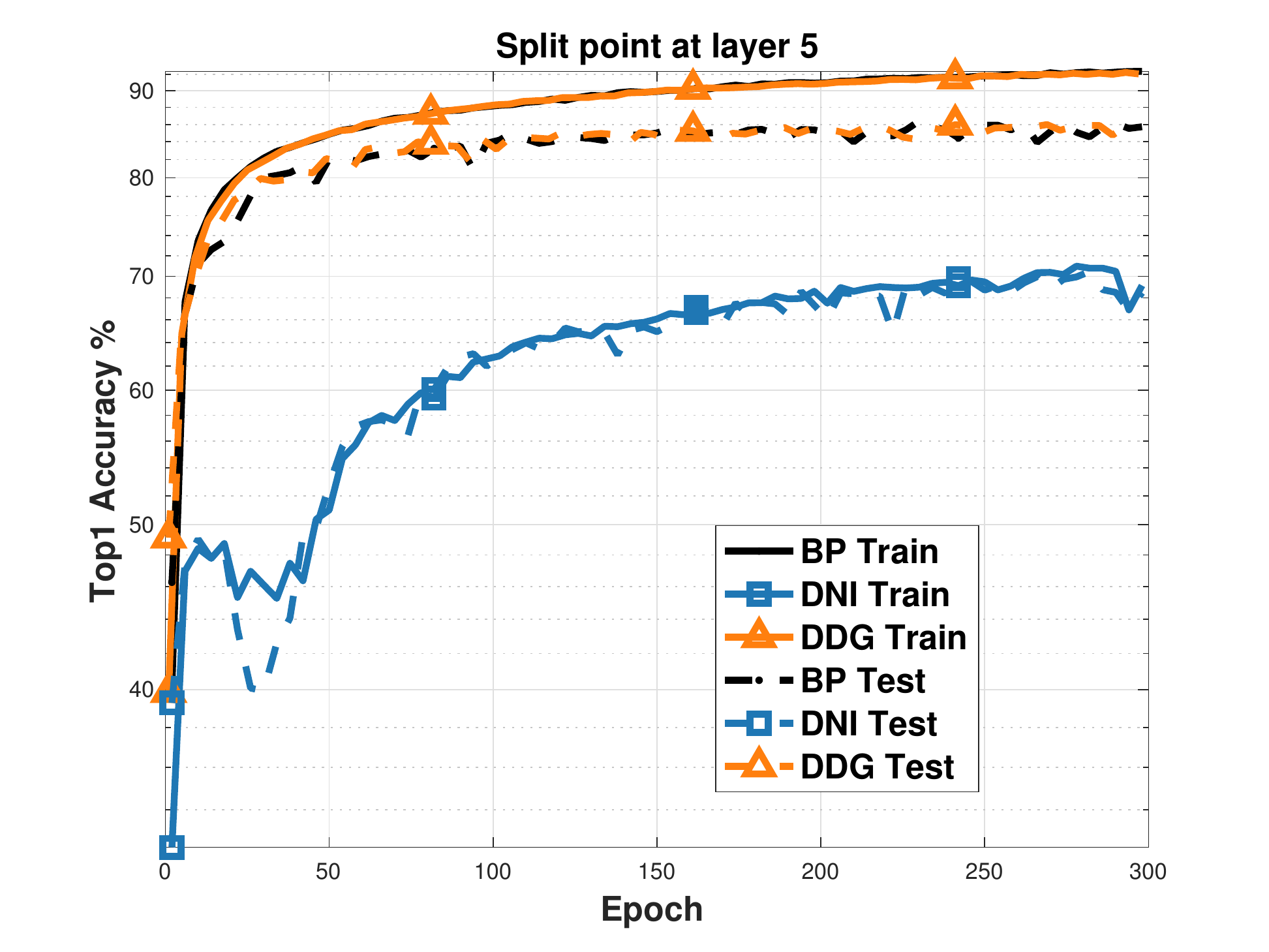}
	\end{subfigure}
	\begin{subfigure}[b]{0.24\textwidth}
		\centering
		\includegraphics[width=1.82in]{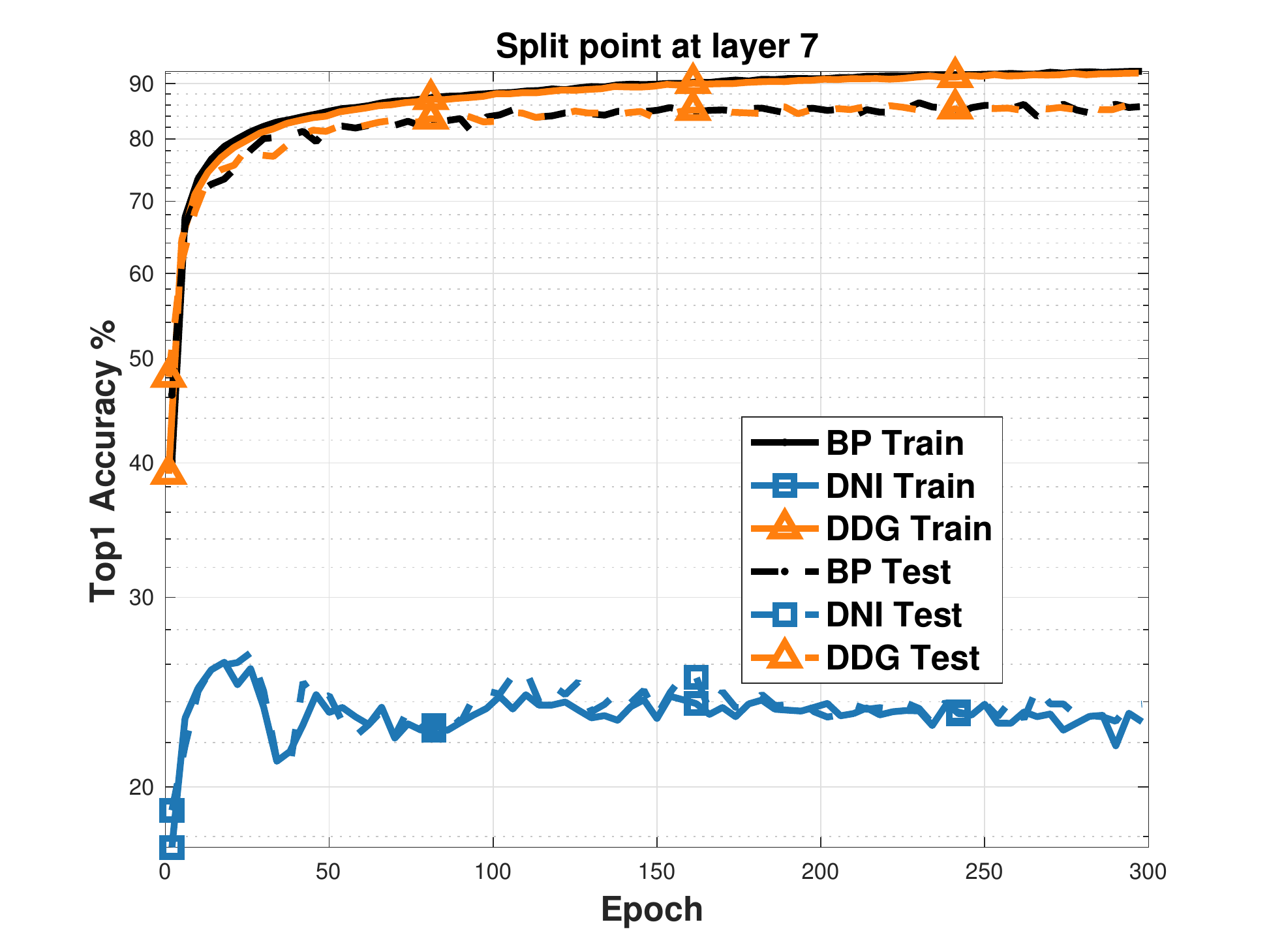}
	\end{subfigure}
	\caption{Training and testing curves regarding epochs for ResNet-8 on CIFAR-10. \textbf{Upper:} Loss function values regarding epochs; \textbf{Bottom:} Top 1 classification accuracies regarding epochs. We split the network into two modules such that there is only one split point in the network for DNI and DDG. }
	\label{odp}
\end{figure*}
Under Assumption \ref{lips} and \ref{bg}, we obtain the following lemma about the sequence of objective functions.
\begin{lemma}
	Assume Assumption \ref{lips} and \ref{bg} hold. In addition, we let $\sigma \coloneqq  \max_t \frac{\gamma_{\max\{0, t-K+1\}}}{\gamma_t}$ and $M_K =KM +  \sigma K^4 M$.  The iterations in Algorithm \ref{alg} satisfy the following inequality, for all $t \in \mathbb{N}$:
	\begin{eqnarray}
	\mathbb{E} \left[ f(w^{t+1}) \right]  - f(w^t)
	\leq -\frac{\gamma_t}{2}  \left\| \nabla f(w^t) \right\|_2^2  +  \gamma_t^2 L M_K
	\label{lem1_iq2}
	\end{eqnarray}
	\label{lem1}
\end{lemma}
From Lemma \ref{lem1}, we can observe that the expected decrease of the objective function is controlled by the stepsize $\gamma_t$ and $M_K$. Therefore, we can guarantee that the values of objective functions are decreasing as long as the stepsizes $\gamma_t$ are small enough such that the right-hand side of (\ref{lem1_iq2}) is less than zero.  Using the lemma above, we can analyze the convergence property for Algorithm \ref{alg}.

\subsection{Fixed Stepsize $\gamma_t$}
Firstly, we analyze the convergence for Algorithm \ref{alg} when $\gamma_t$ is fixed and prove that the learned model will converge sub-linearly to the neighborhood of the critical points.
\begin{theorem}
		\label{them1}
	Assume Assumption \ref{lips} and \ref{bg} hold and the fixed stepsize sequence $\{ \gamma_t\}$ satisfies $\gamma_t = \gamma$ and $\gamma L \leq 1 , \forall t \in \{0,1,...,T-1\}$.  In addition, we assume $w^*$ to be the optimal solution to $f(w)$ and let $\sigma = 1$ such that $M_K =KM +  K^4 M$. Then, the output of Algorithm \ref{alg} satisfies that:
{\small
	\begin{eqnarray}
	\frac{1}{T} \sum\limits_{t=0}^{T-1}\mathbb{E}  \left\| \nabla f(w^t) \right\|_2^2 \leq \frac{2\left( f(w^0) - f(w^*) \right)}{\gamma T} + {2\gamma L M_K}
\end{eqnarray} }
\end{theorem}
In Theorem \ref{them1}, we can observe that when $T \rightarrow \infty$, the average norm of the gradients is upper bounded by $2\gamma L M_K$. The number of modules $K$ affects the value of the upper bound. 
Selecting a small stepsize $\gamma$ allows us to get better neighborhood to the critical points,  however it also seriously decreases the speed of convergence. 

\subsection{Diminishing Stepsize $\gamma_t$}
In this section, we prove that Algorithm \ref{alg} with diminishing stepsizes can guarantee the convergence to critical points for the non-convex problem.
\begin{theorem}
	Assume Assumption \ref{lips} and \ref{bg} hold and the diminishing stepsize sequence $\{ \gamma_t\}$ satisfies $\gamma_t = \frac{\gamma_0}{1+t}$ and $\gamma_t L \leq 1 , \forall t \in \{0,1,...,T-1\}$.  In addition, we assume $w^*$ to be the optimal solution to $f(w)$ and let $\sigma = K$ such that $M_K =KM +  K^5 M$. Setting $\Gamma_T = \sum\limits_{t=0}^{T-1} \gamma_t $, then the output of Algorithm \ref{alg} satisfies that:
\begin{eqnarray}
\frac{1}{\Gamma_T} \sum\limits_{t=0}^{T-1} \gamma_t\mathbb{E} \left\| \nabla f(w^t) \right\|_2^2 &\leq& \frac{2\left( f(w^0) - f(w^*) \right)}{\Gamma_T} \nonumber \\
&& + \frac{2\sum\limits_{t=0}^{T-1} \gamma_t^2 L M_K}{\Gamma_T}
\label{them2_iq_1001}
\end{eqnarray} 
	\label{them2}
\end{theorem}
\begin{corollary}
Since $\gamma_t = \frac{\gamma_0}{t+1}$, the stepsize requirements in \cite{robbins1951stochastic} are satisfied that:
\begin{eqnarray}
\lim_{T\rightarrow \infty} \sum\limits_{t=0}^{T-1} \gamma_t = \infty \hspace{0.3cm} \text{and} \hspace{0.3cm} 
   \lim_{T\rightarrow \infty} \sum\limits_{t=0}^{T-1} \gamma_t^2 < \infty.
\end{eqnarray} 
Therefore, according to Theorem \ref{them2}, when $T\rightarrow \infty$, the right-hand side of (\ref{them2_iq_1001}) converges to $0$. 
\end{corollary}
\begin{corollary}
Suppose $w^s$ is chosen randomly from $\{w^t \}_{t=0}^{T-1}$  with probabilities proportional to  $ \{\gamma_t \}_{t=0}^{T-1}$.  According to  Theorem \ref{them2}, we can prove that Algorithm \ref{alg} guarantees convergence to critical points for the non-convex problem:
\begin{eqnarray}
\lim\limits_{s\rightarrow \infty} \mathbb{E} \|\nabla f(w^s)\|_2^2& =& 0
\end{eqnarray}
\end{corollary}

\begin{figure*}[th]
	\centering
	\begin{subfigure}[b]{0.33\textwidth}
		\centering
		\includegraphics[width=2.4in]{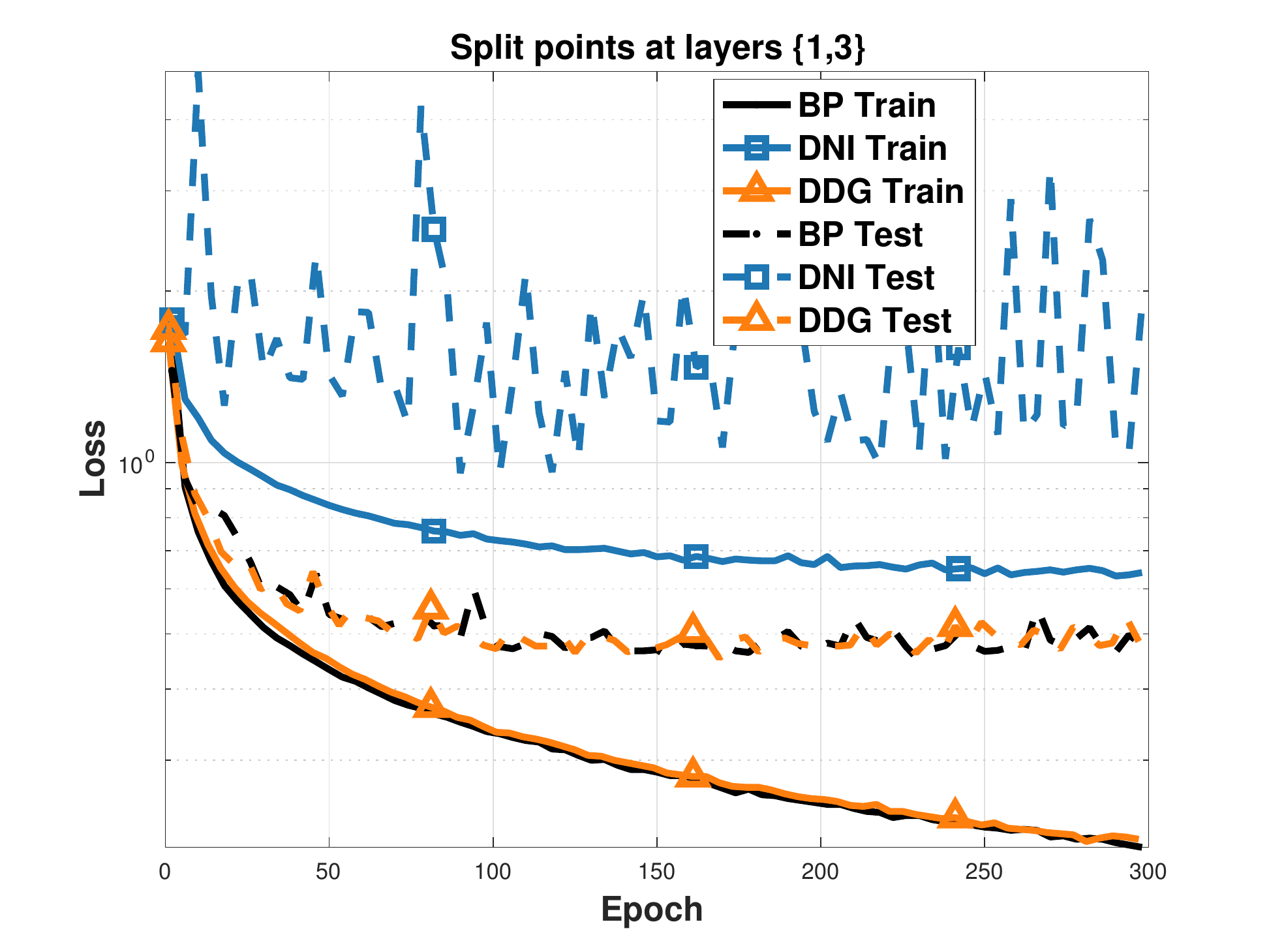}
	\end{subfigure}
	\begin{subfigure}[b]{0.33\textwidth}
		\centering
		\includegraphics[width=2.4in]{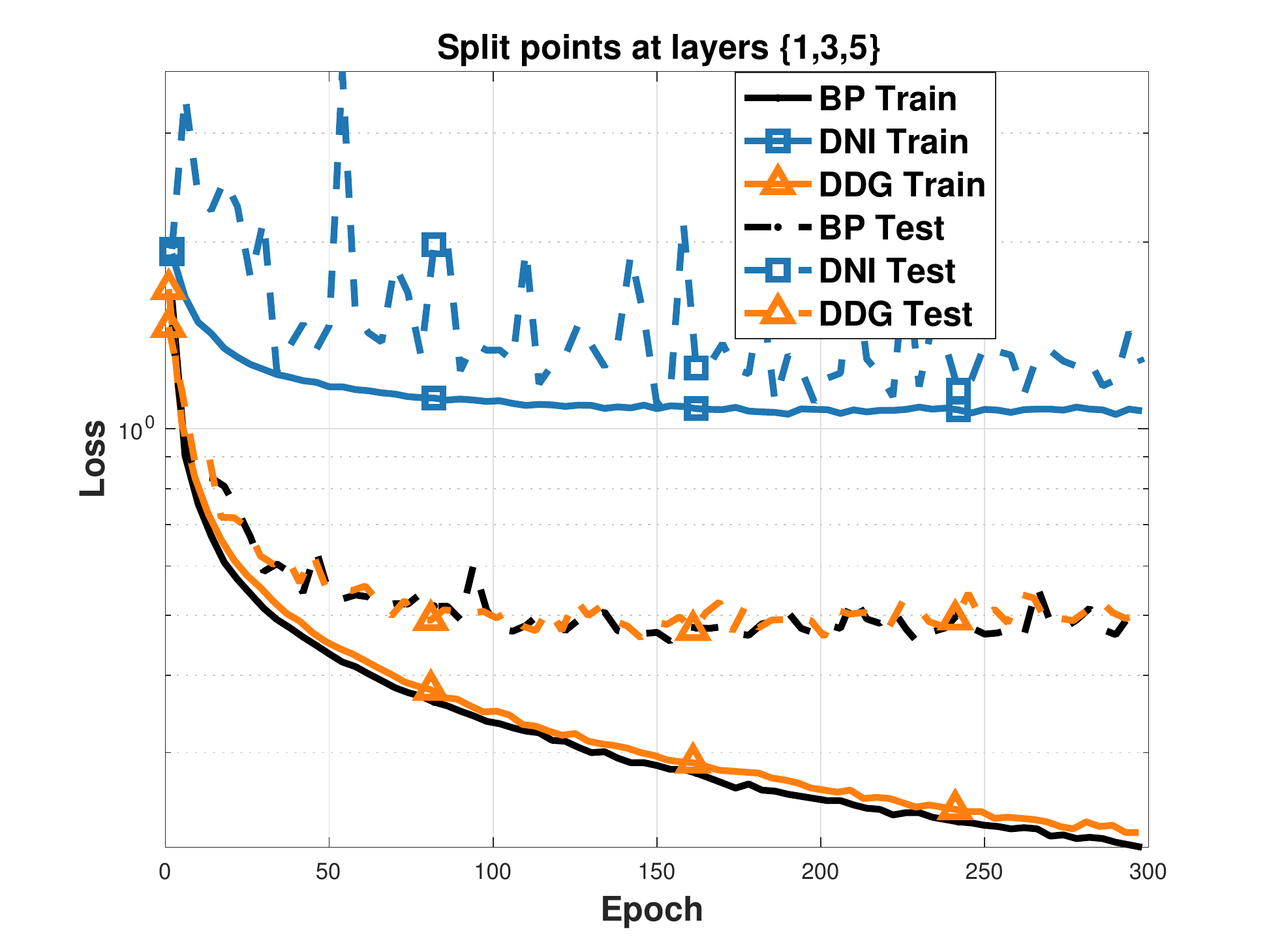}
	\end{subfigure}
	\begin{subfigure}[b]{0.33\textwidth}
		\centering
		\includegraphics[width=2.4in]{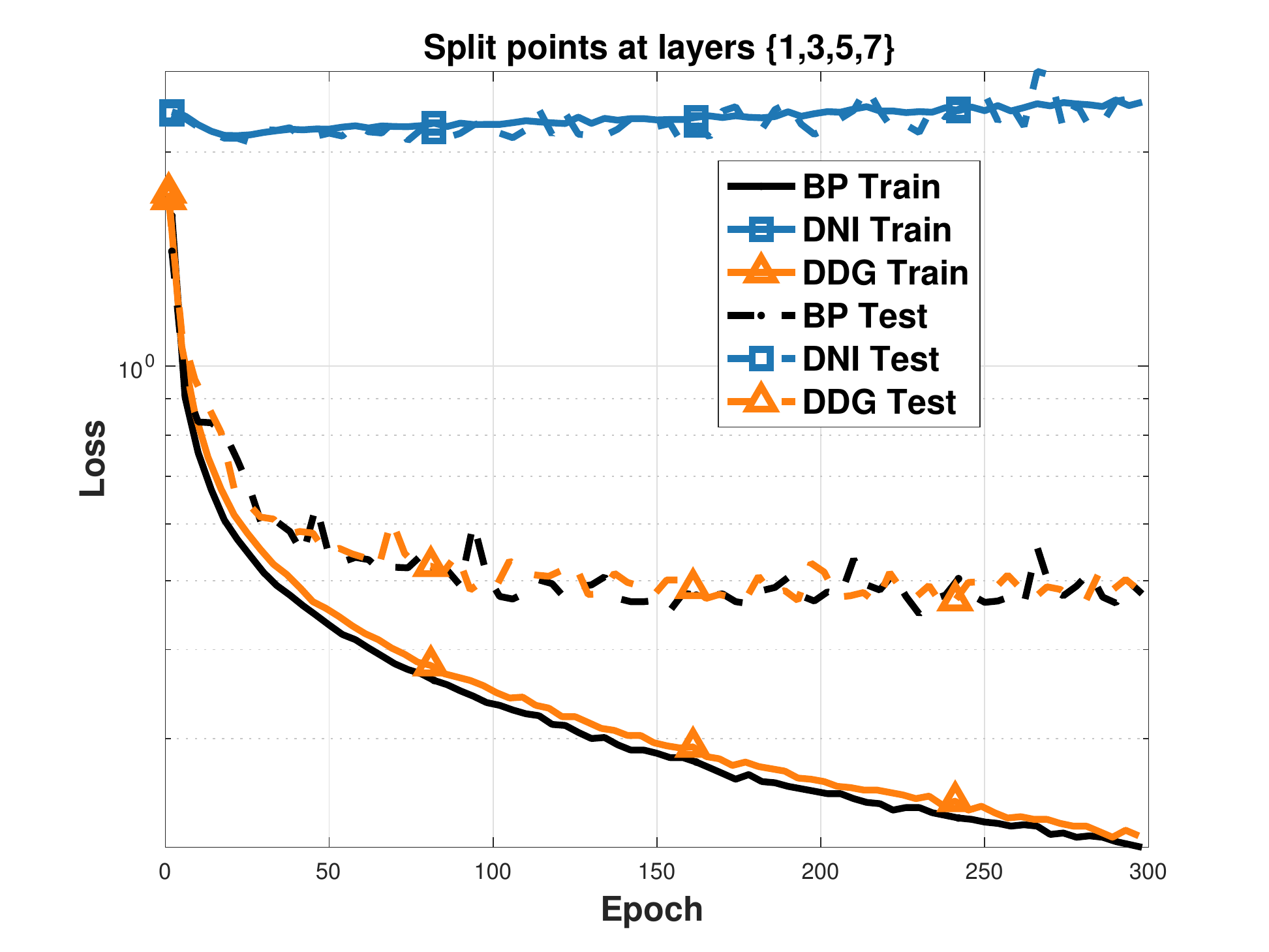}
	\end{subfigure}
	\begin{subfigure}[b]{0.33\textwidth}
		\centering
		\includegraphics[width=2.4in]{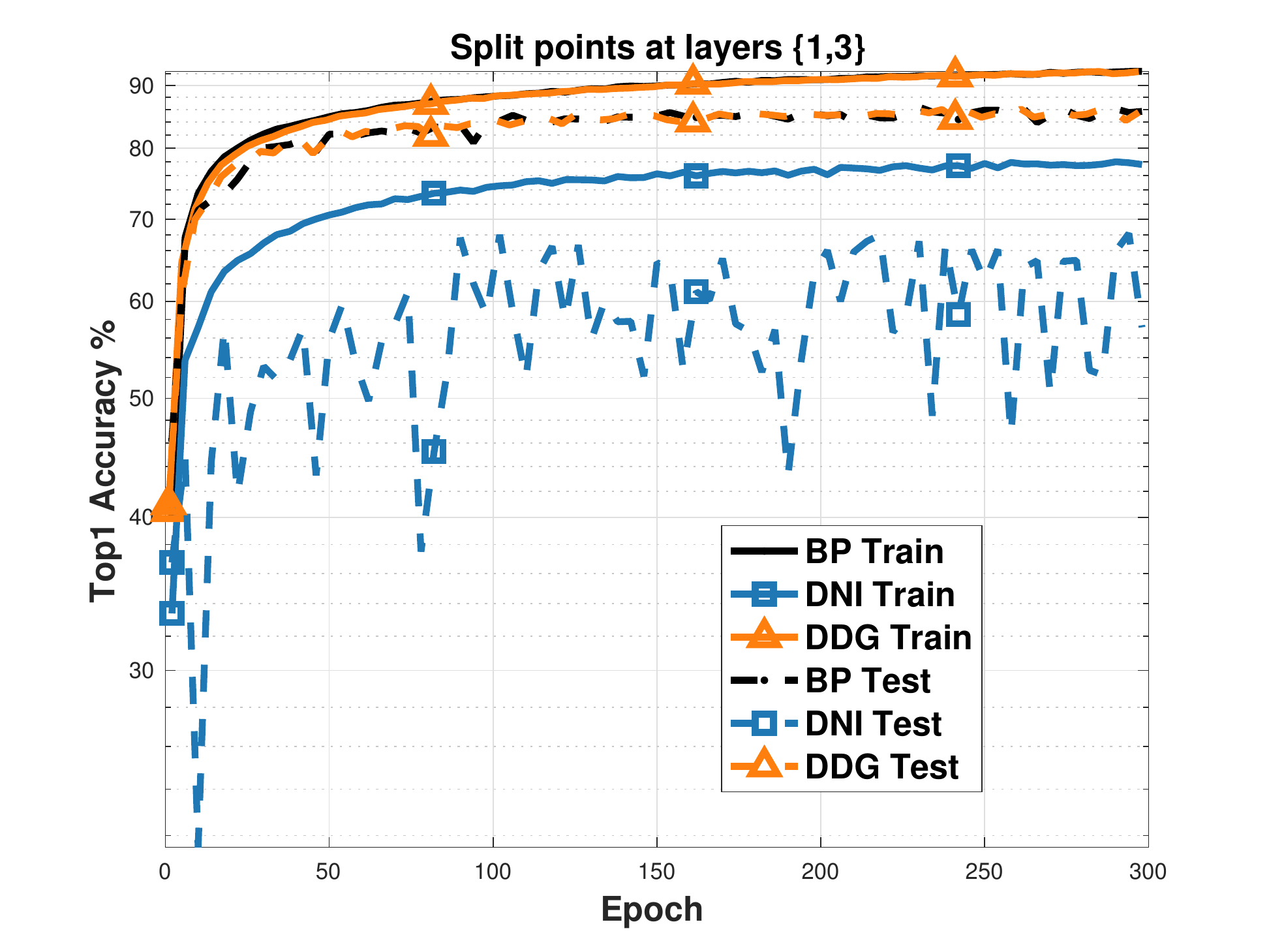}
	\end{subfigure} 
	\begin{subfigure}[b]{0.33\textwidth}
		\centering
		\includegraphics[width=2.4in]{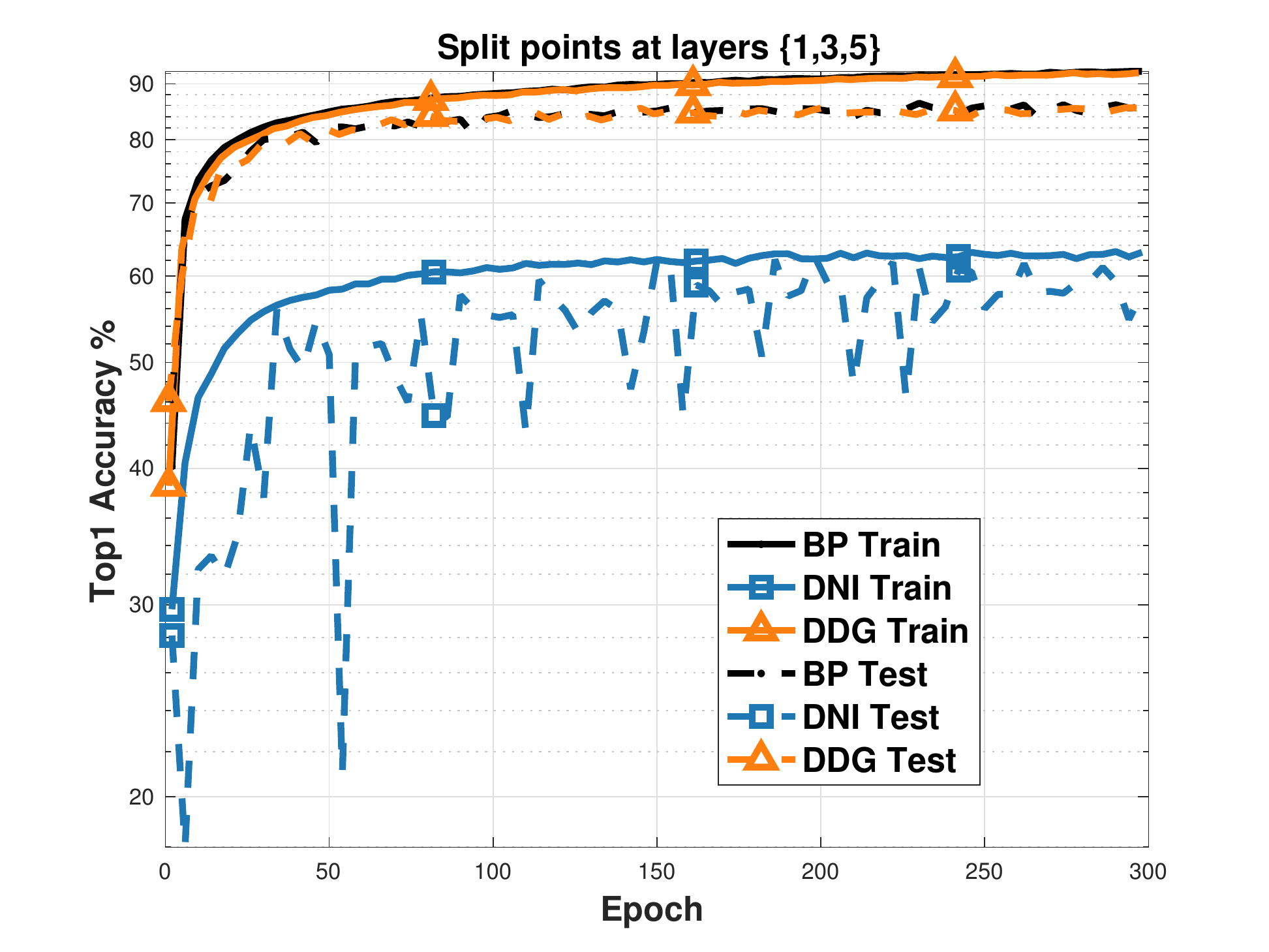}
	\end{subfigure}
	\begin{subfigure}[b]{0.33\textwidth}
		\centering
		\includegraphics[width=2.4in]{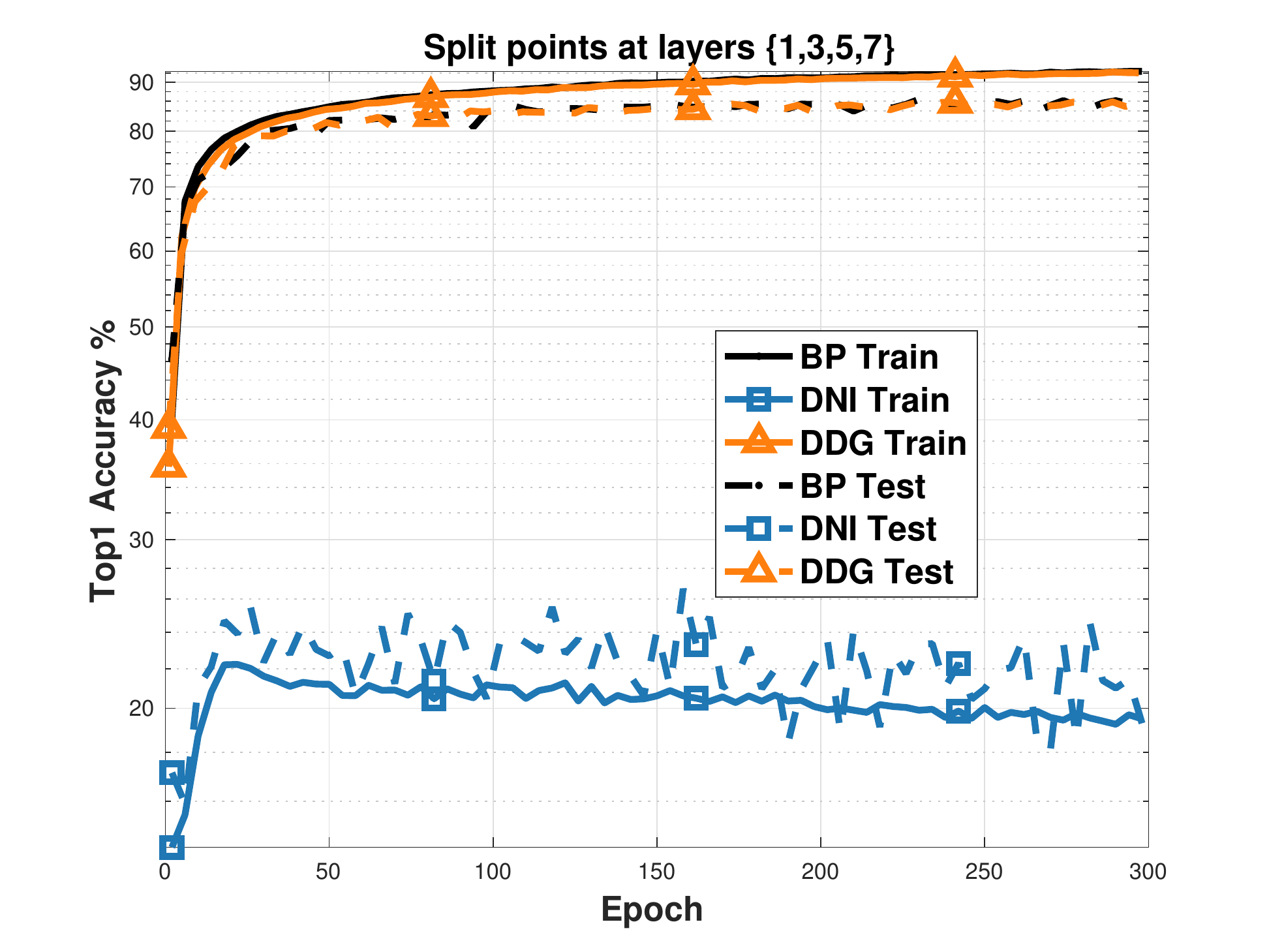}
	\end{subfigure}
	\caption{Training and testing curves regarding epochs for ResNet-8 on CIFAR-10. \textbf{Upper:} Loss function values regarding epochs; \textbf{Bottom:} Top1 classification accuracies regarding epochs. For DNI and DDG, the number of split points in the network ranges from $2$ to $4$.}
	\label{mdp}
\end{figure*}

\begin{figure*}[h]
	\centering
	\begin{subfigure}[b]{0.43\textwidth}
		\centering
		\includegraphics[width=2.9in]{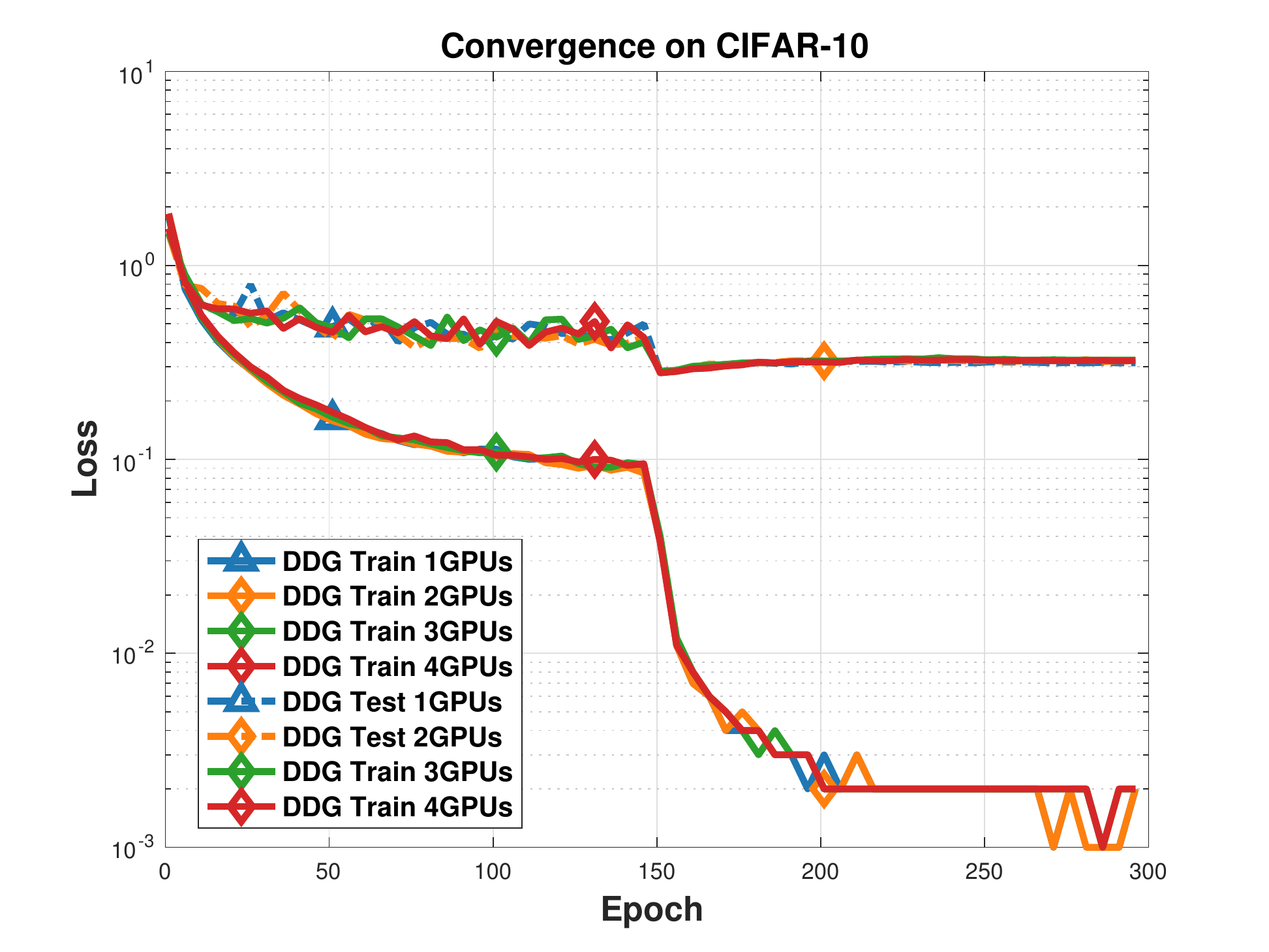}
		\caption{}
		\label{dist0}
	\end{subfigure}
	\begin{subfigure}[b]{0.43\textwidth}
	\centering
	\includegraphics[width=2.9in]{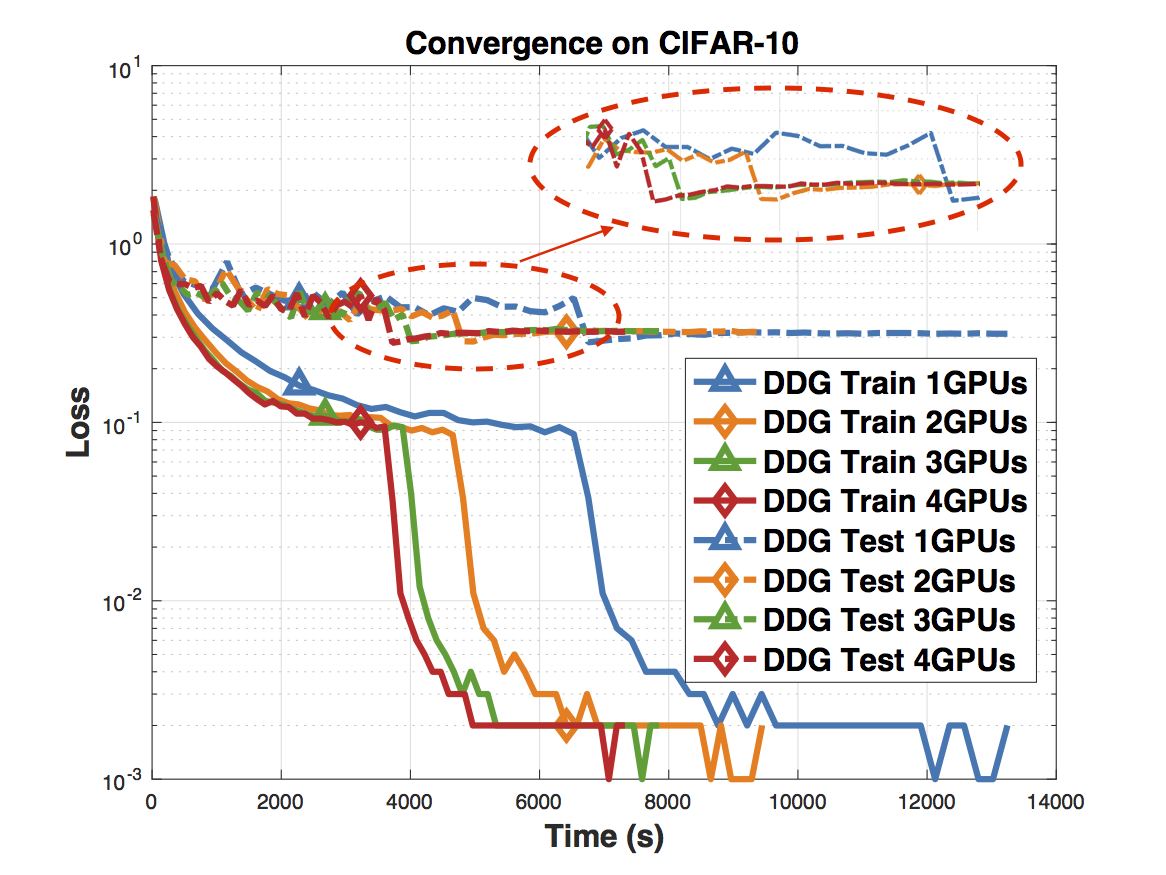}
			\caption{}
	\label{dist1}
\end{subfigure}
\caption{ Training and testing loss curves for ResNet-110 on CIFAR-10 using multiple GPUs. \textbf{(\ref{dist0})} Loss function value regarding epochs. \textbf{(\ref{dist1})}  Loss function value regarding computation time.  
}
\label{dist_conv}
\end{figure*}

\begin{figure}[h]
	\centering
		\centering
		\includegraphics[width=2.8in]{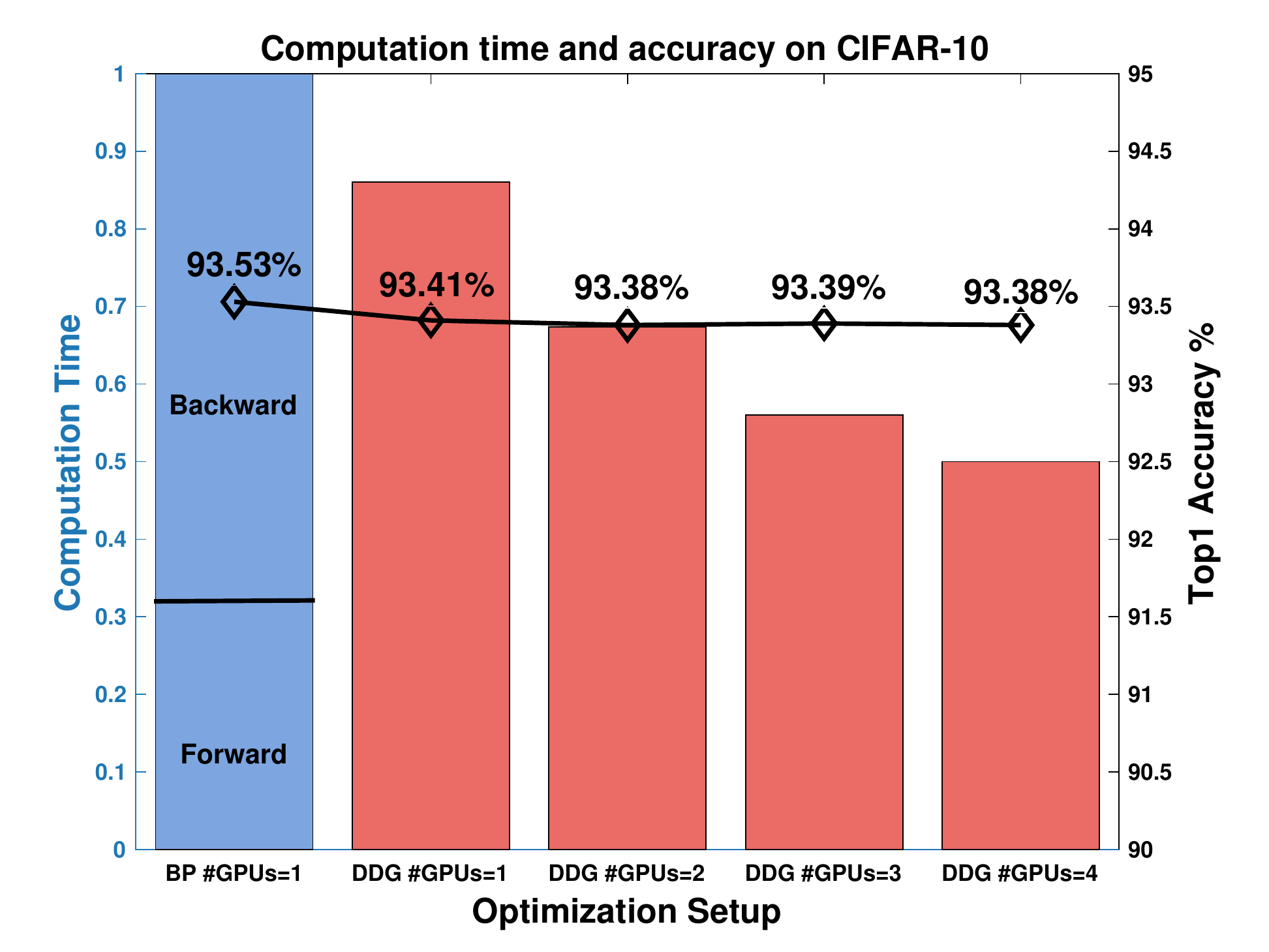}
	\caption{Computation time and the best Top 1 accuracy for ResNet-110 on the test data of CIFAR-10. The most left bar denotes the computation time using backpropagation algorithm on a GPU, where the forward time accounts for about $32\%$.  We normalize the computation time of all optimization settings using the amount of time required by backpropagation.
	}
		\label{dist2}
\end{figure}

\section{Experiments}
\label{sec_exp}
In this section, we experiment with ResNet \cite{he2016deep} on image classification benchmark datasets: CIFAR-10 and CIFAR-100 \cite{krizhevsky2009learning}. In section \ref{exp_classification}, we evaluate our method by varying the positions and the number of the split points in the network; In section \ref{exp_deep} we use our method to optimize deeper neural networks and show that its performance is as good as the performance of backpropagation;  finally, we split and distribute the ResNet-110 across GPUs in Section \ref{distributed_training}, results showing that the proposed method achieves a speedup of two times without loss of accuracy.

 \textbf{Implementation Details:}  We implement DDG algorithm using PyTorch library \cite{paszke2017automatic}. The trained network is split into $K$ modules where each module is running on a subprocess. 
 The subprocesses are spawned using multiprocessing package \footnote{https://docs.python.org/3/library/multiprocessing.html$\#$module-multiprocessing} such that we can fully leverage multiple processors on a given machine. Running modules on different subprocesses make the communication very difficult. To make the communication fast, we utilize the shared memory objects in the multiprocessing package. As in Figure \ref{algo}, every two adjacent modules share a pair of activation ($h$) and error gradient ($\delta$).

 \subsection{Comparison of BP, DNI and DDG}
  \label{exp_classification}
In this section, we train ResNet-8 on CIFAR-10 on a single Titan X GPU. The architecture of the ResNet-8 is in Table \ref{architecture}. All experiments are run for $300$ epochs and optimized using Adam  optimizer \cite{kingma2014adam} with a batch size of $128$. The stepsize is initialized at $1\times 10^{-3}$. We augment the dataset with random cropping, random horizontal flipping and normalize the image using mean and standard deviation. There are three compared methods in this experiment:
\begin{itemize}
	\item \textbf{BP}:  Adam optimizer in Pytorch uses backpropagation algorithm with data parallelism \cite{rumelhart1988learning} to compute gradients. 	 
	\item \textbf{DNI}:  Decoupled neural interface (DNI)  in \cite{jaderberg2016decoupled}. Following \cite{jaderberg2016decoupled}, the synthetic network is a stack of  three convolutional layers with $L\hspace{0.1cm} 5\times5$ filters with resolution preserving padding. The filter depth $L$ is determined by the position of DNI.  We also input label information into the synthetic network to increase  final accuracy. 
	\item \textbf{DDG}: Adam optimizer using delayed gradients in Algorithm \ref{alg2}.
\end{itemize}

 \begin{table}[t]
	\caption{Architectural details. \textbf{Units} denotes the number of residual units in each group. Each unit is a basic residual block without bottleneck.  \textbf{Channels} indicates the number of filters used in each unit in each group.} 
		\center
	\begin{tabular}{c|c|c}
		\hline
		\textbf{	Architecture} &  \textbf{Units}	 &  \textbf{Channels}  \\ 
		\hline 
		ResNet-8 & 1-1-1 & 16-16-32-64\\  \hline
		ResNet-56 & 9-9-9 & 16-16-32-64\\  \hline
		ResNet-110 & 18-18-18& 16-16-32-64  \\ \hline
	\end{tabular}
	\label{architecture}
\end{table}
\textbf{Impact of split position (depth).} 
The position (depth) of the split points determines the number of layers using delayed gradients.  Stale or synthetic gradients will induce noises in the training process, affecting the convergence of the objective.
 Figure \ref{odp} exhibits the experimental results when there is only one split point with varying positions. 
 In the first column, we know that all compared methods have similar performances when the split point is at layer $1$. DDG performs consistently well when we place the split point at deeper positions $3,5$ or $7$. On the contrary, the performance of DNI degrades as we vary the positions and it cannot even converge when the split point is at layer $7$.

\textbf{Impact of the number of split points.} From equation (\ref{backward4}), we know that the maximum time delay is determined by the number of modules $K$.  Theorem \ref{them2} also shows that $K$ affects the convergence rate.
In this experiment, we vary the number of split points in the network from $2$ to $4$ and plot the results in Figure \ref{mdp}. 
It is easy to observe that DDG performs as well as BP, regardless of the number of split points in the network. However, DNI is very unstable when we place more split points, and cannot even converge sometimes.

\begin{figure*}[h]
	\centering
	\begin{subfigure}[b]{0.24\textwidth}
		\centering
		\includegraphics[width=1.82in]{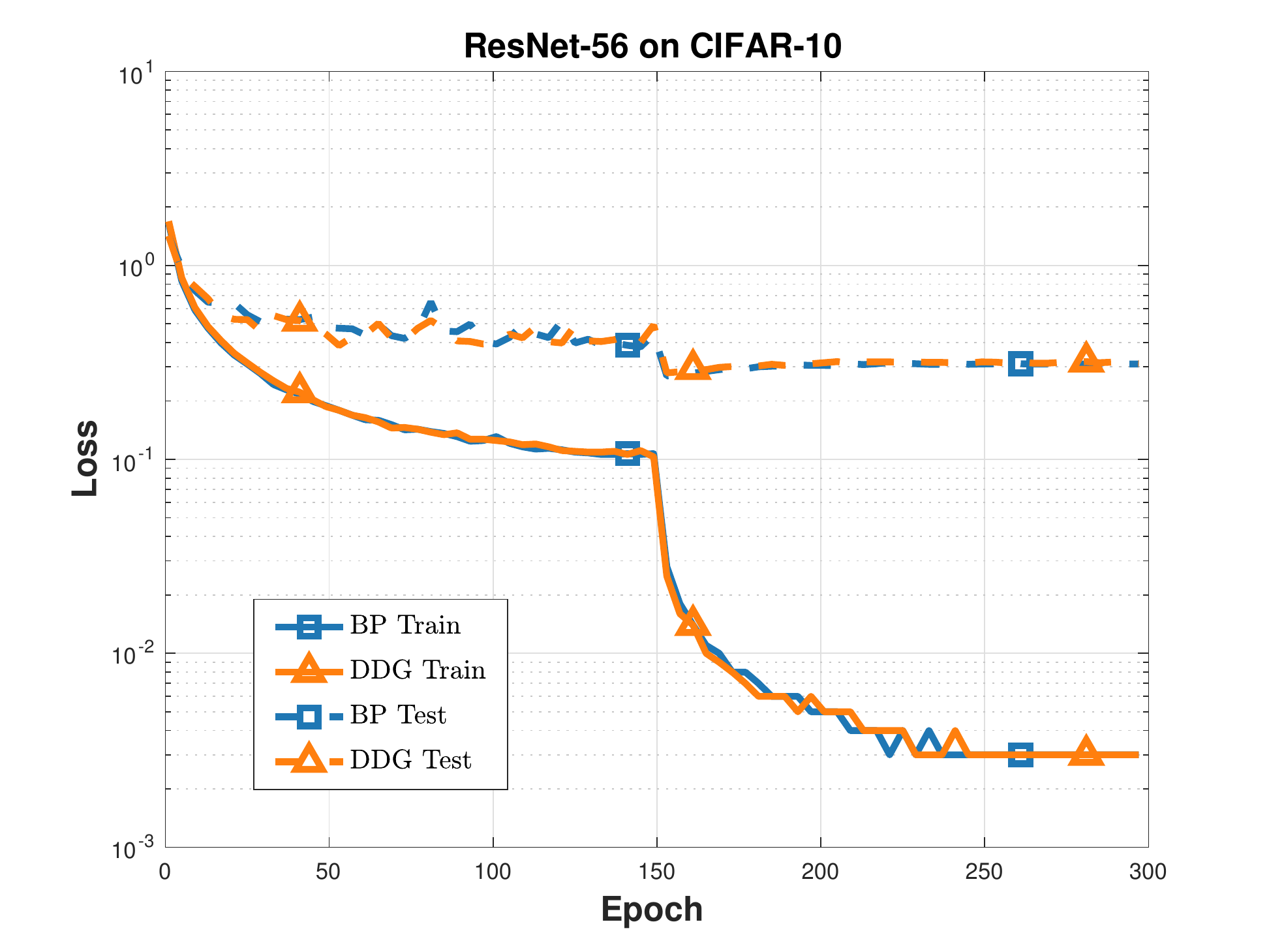}
	\end{subfigure}
	\begin{subfigure}[b]{0.24\textwidth}
		\centering
		\includegraphics[width=1.82in]{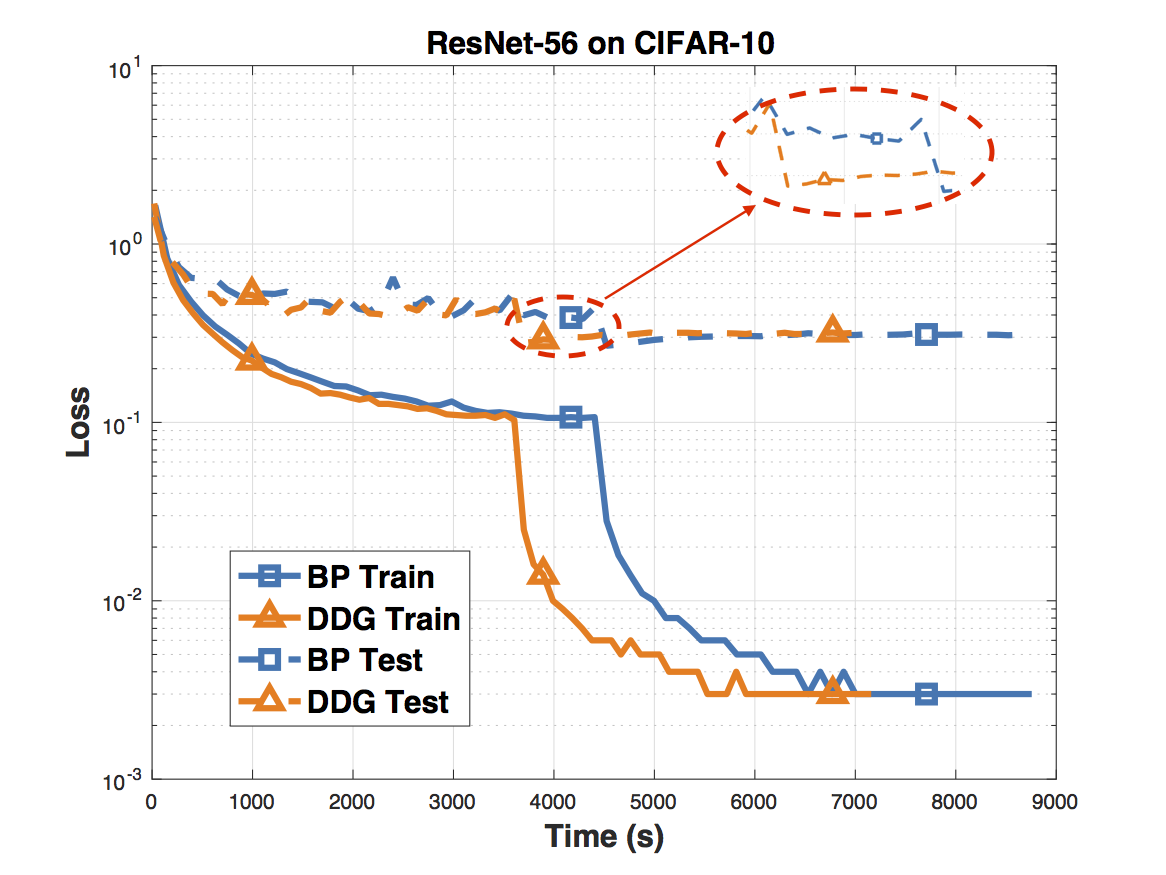}
	\end{subfigure}
	\begin{subfigure}[b]{0.24\textwidth}
		\centering
		\includegraphics[width=1.82in]{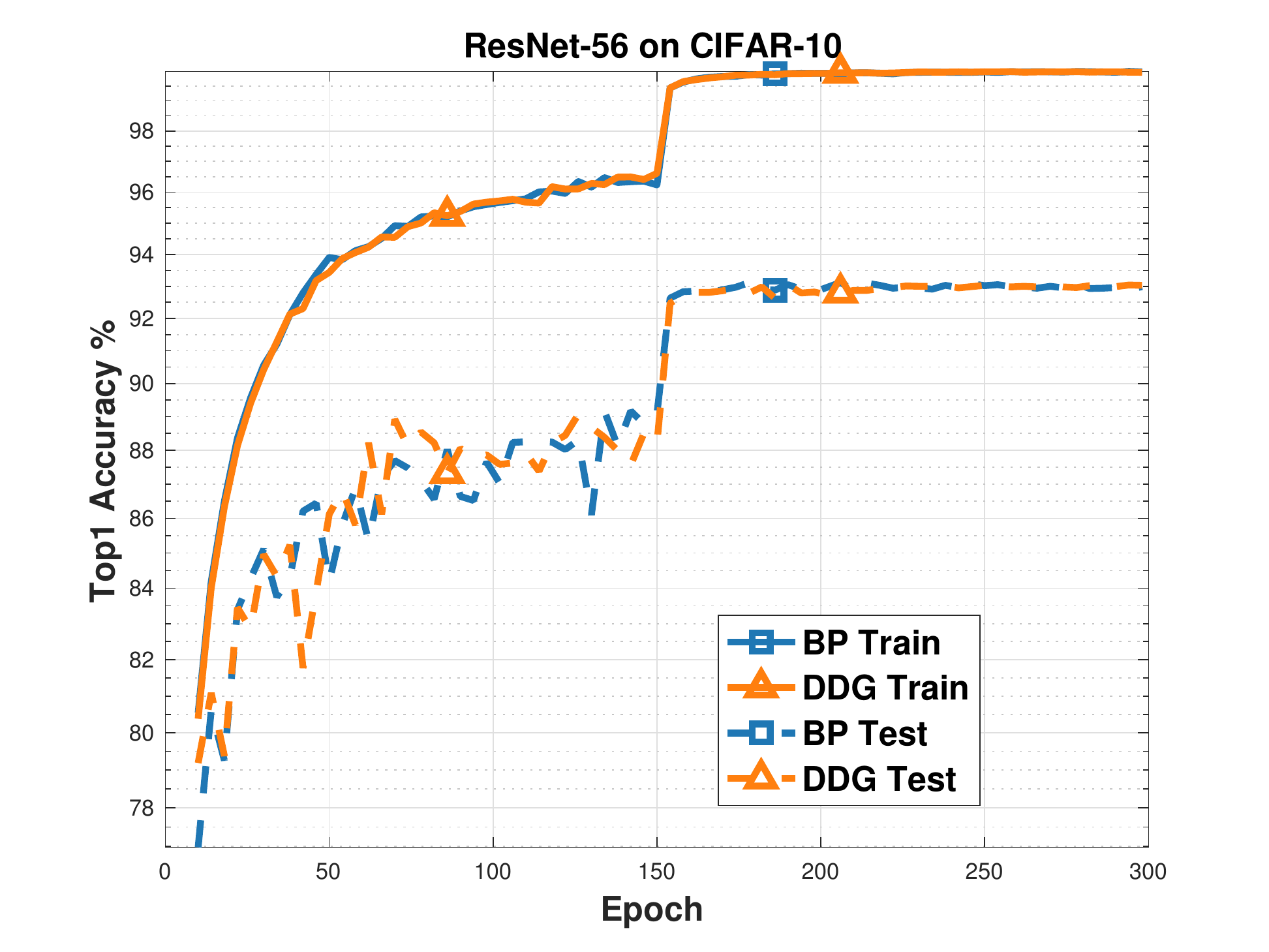}
	\end{subfigure}
	\begin{subfigure}[b]{0.24\textwidth}
		\centering
		\includegraphics[width=1.82in]{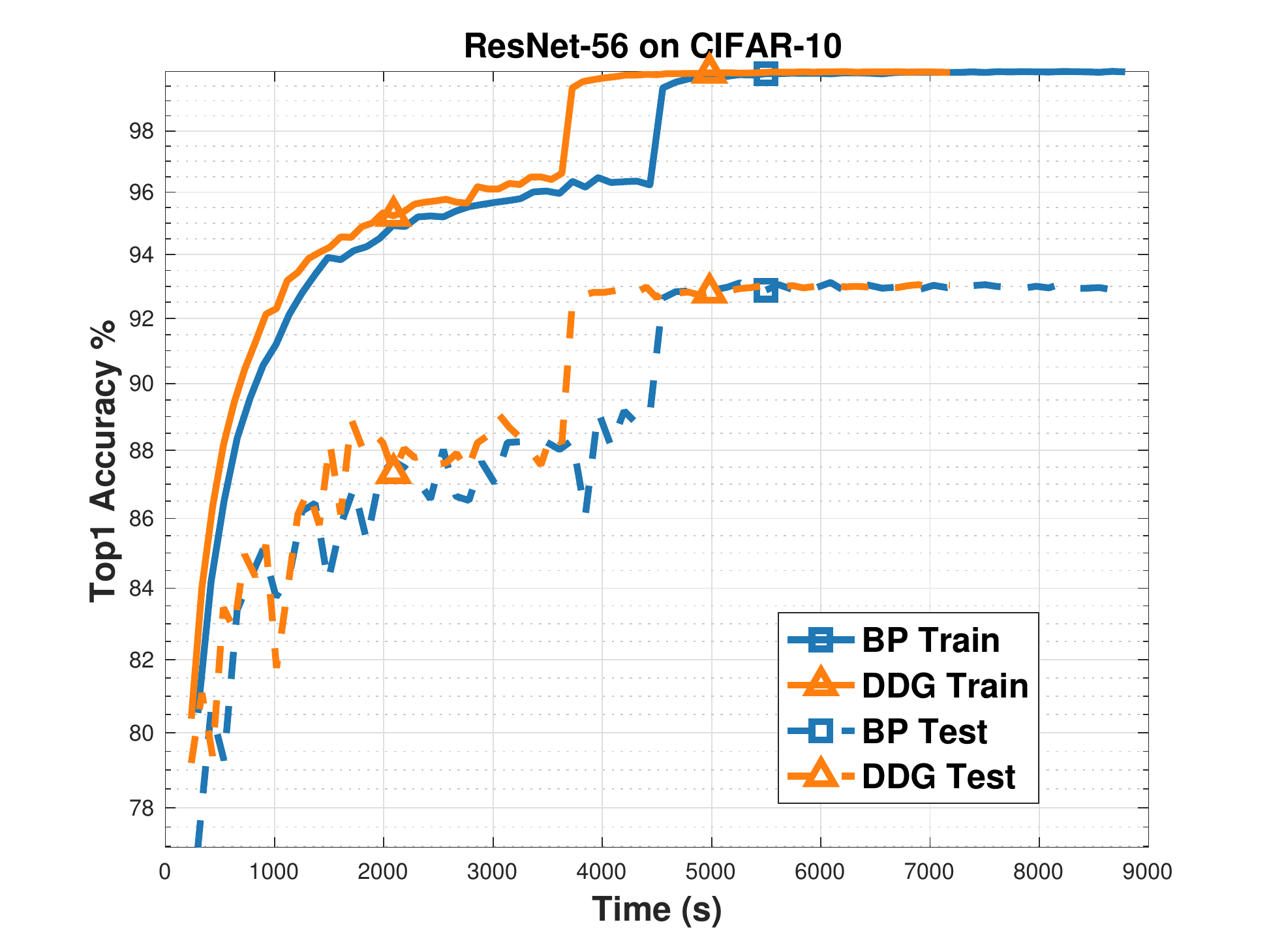}
	\end{subfigure}
	\begin{subfigure}[b]{0.24\textwidth}
		\centering
		\includegraphics[width=1.82in]{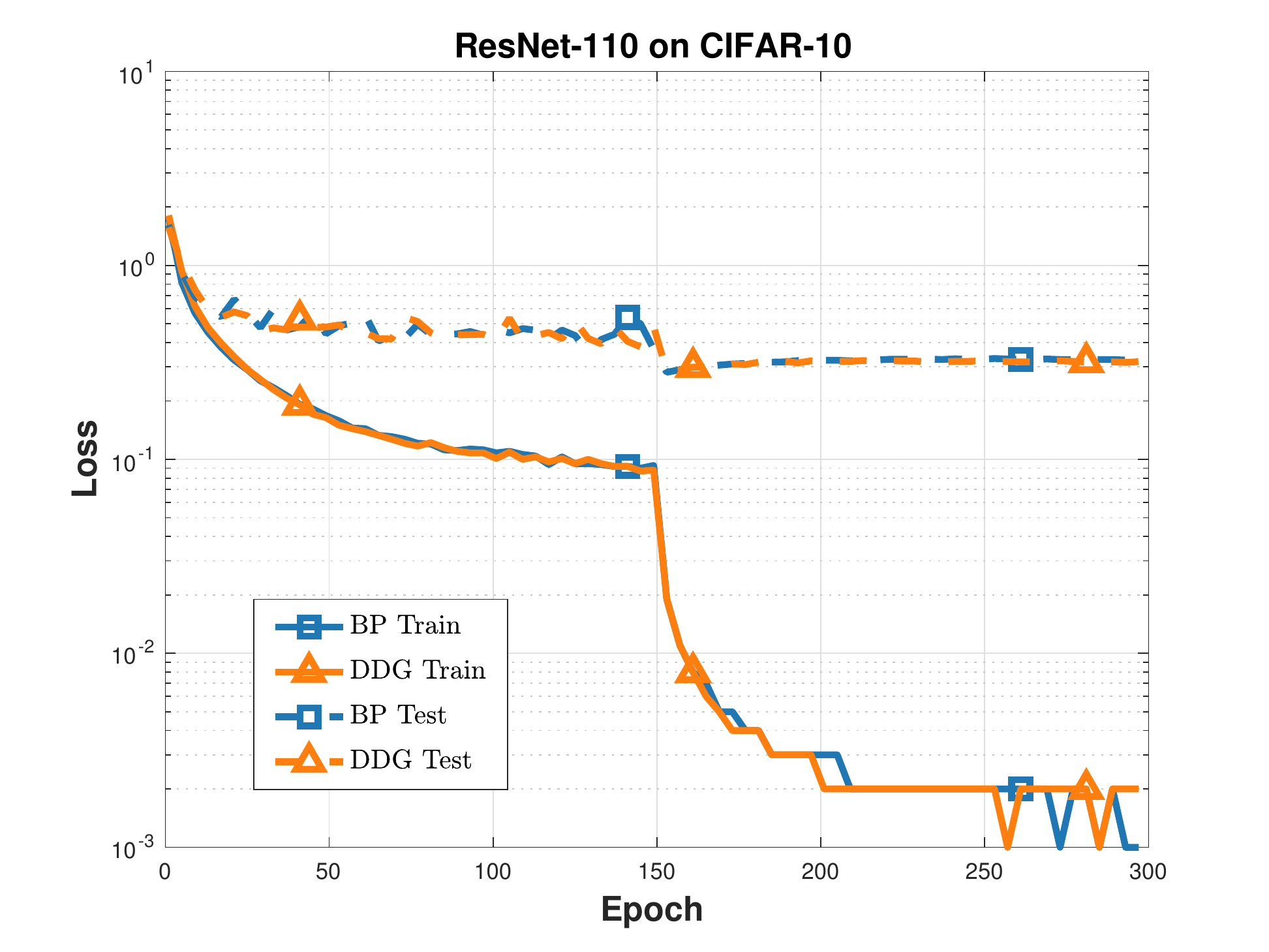}
	\end{subfigure}
	\begin{subfigure}[b]{0.24\textwidth}
		\centering
		\includegraphics[width=1.82in]{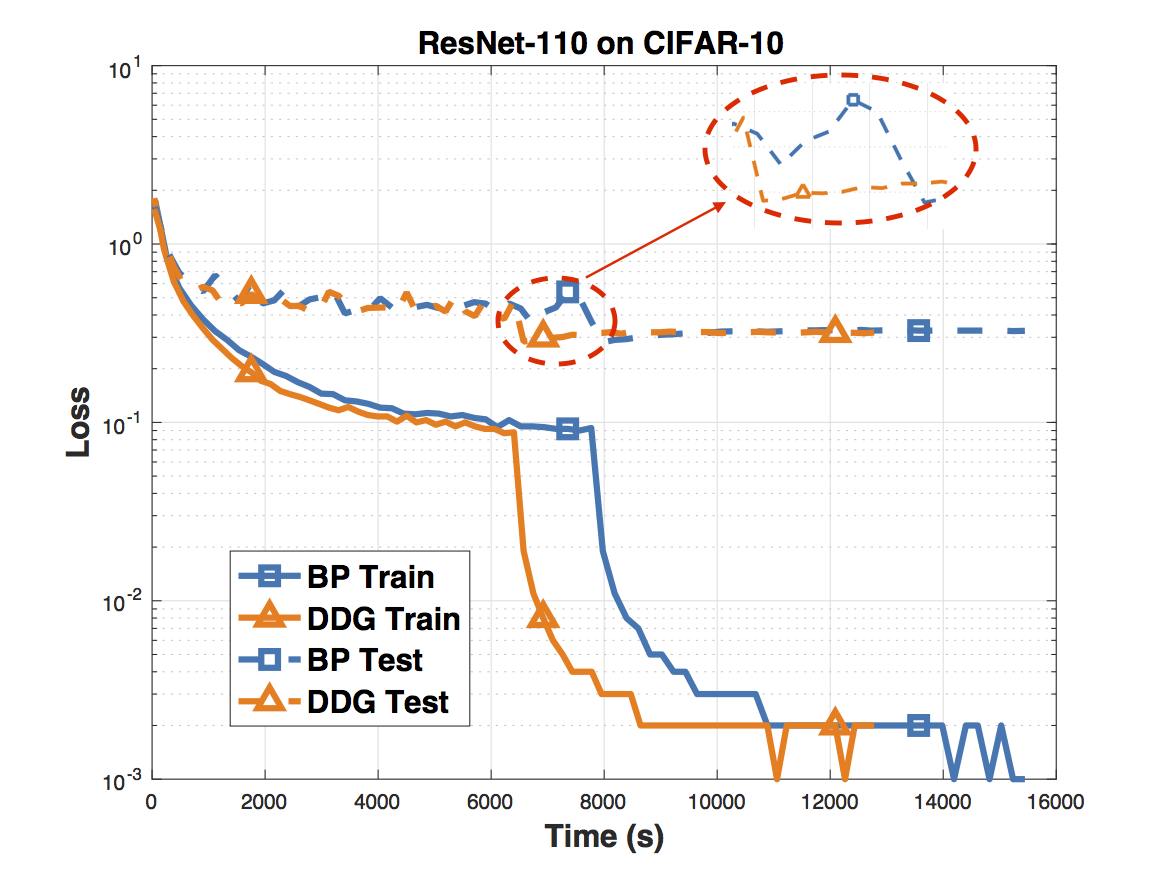}
	\end{subfigure}
	\begin{subfigure}[b]{0.24\textwidth}
		\centering
		\includegraphics[width=1.82in]{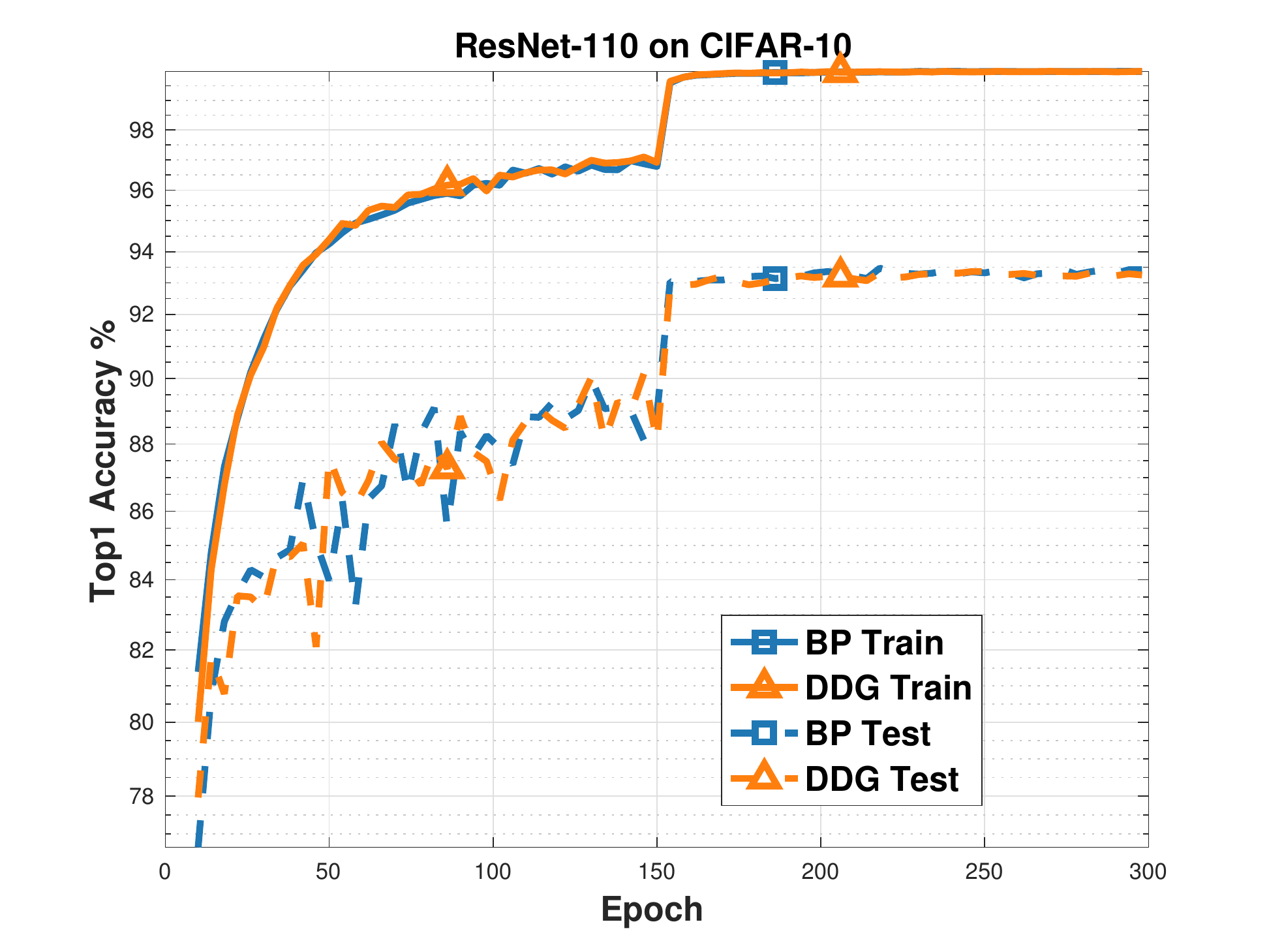}
	\end{subfigure}
	\begin{subfigure}[b]{0.24\textwidth}
		\centering
		\includegraphics[width=1.82in]{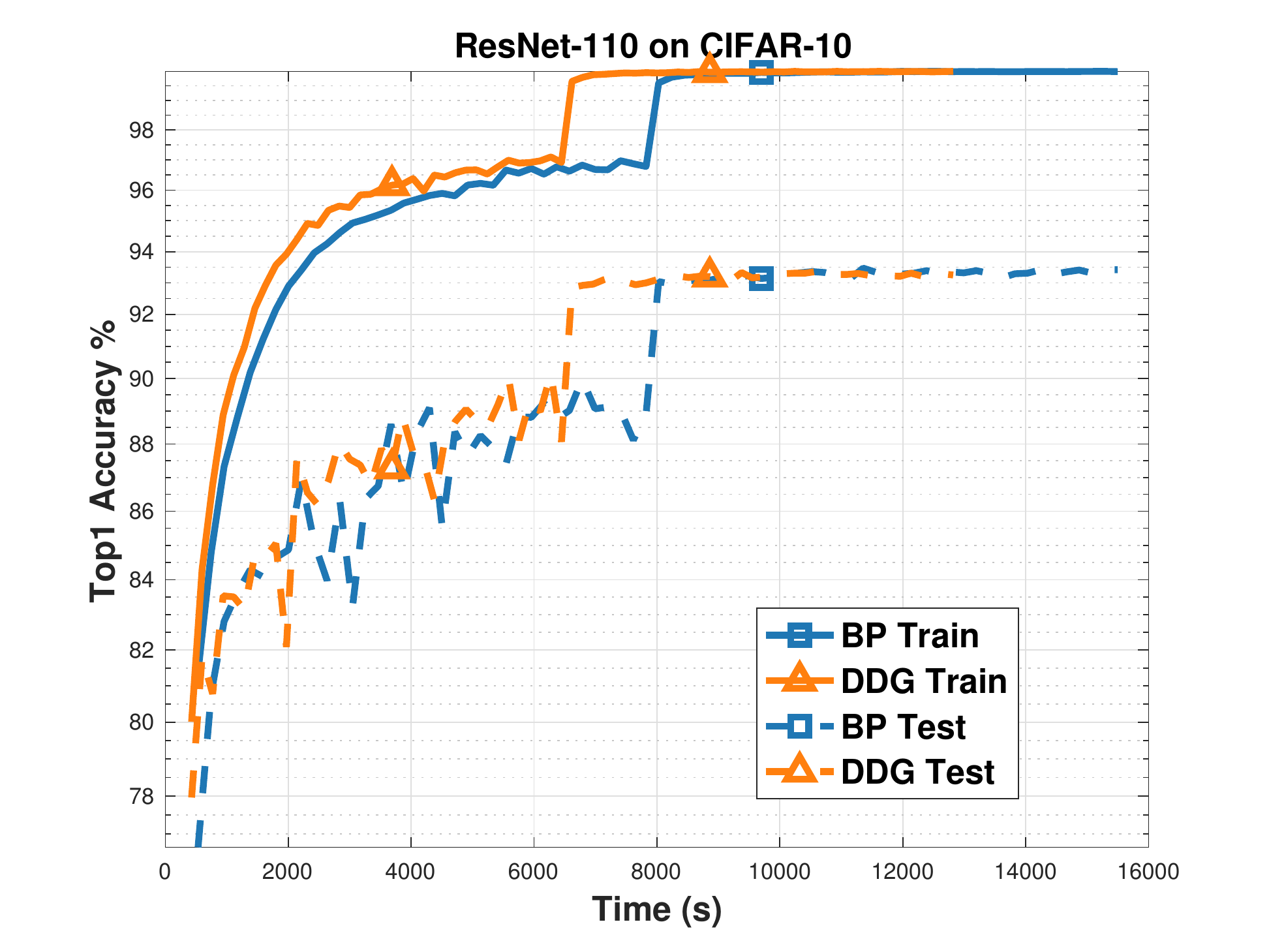}
	\end{subfigure}
	\begin{subfigure}[b]{0.24\textwidth}
		\centering
		\includegraphics[width=1.82in]{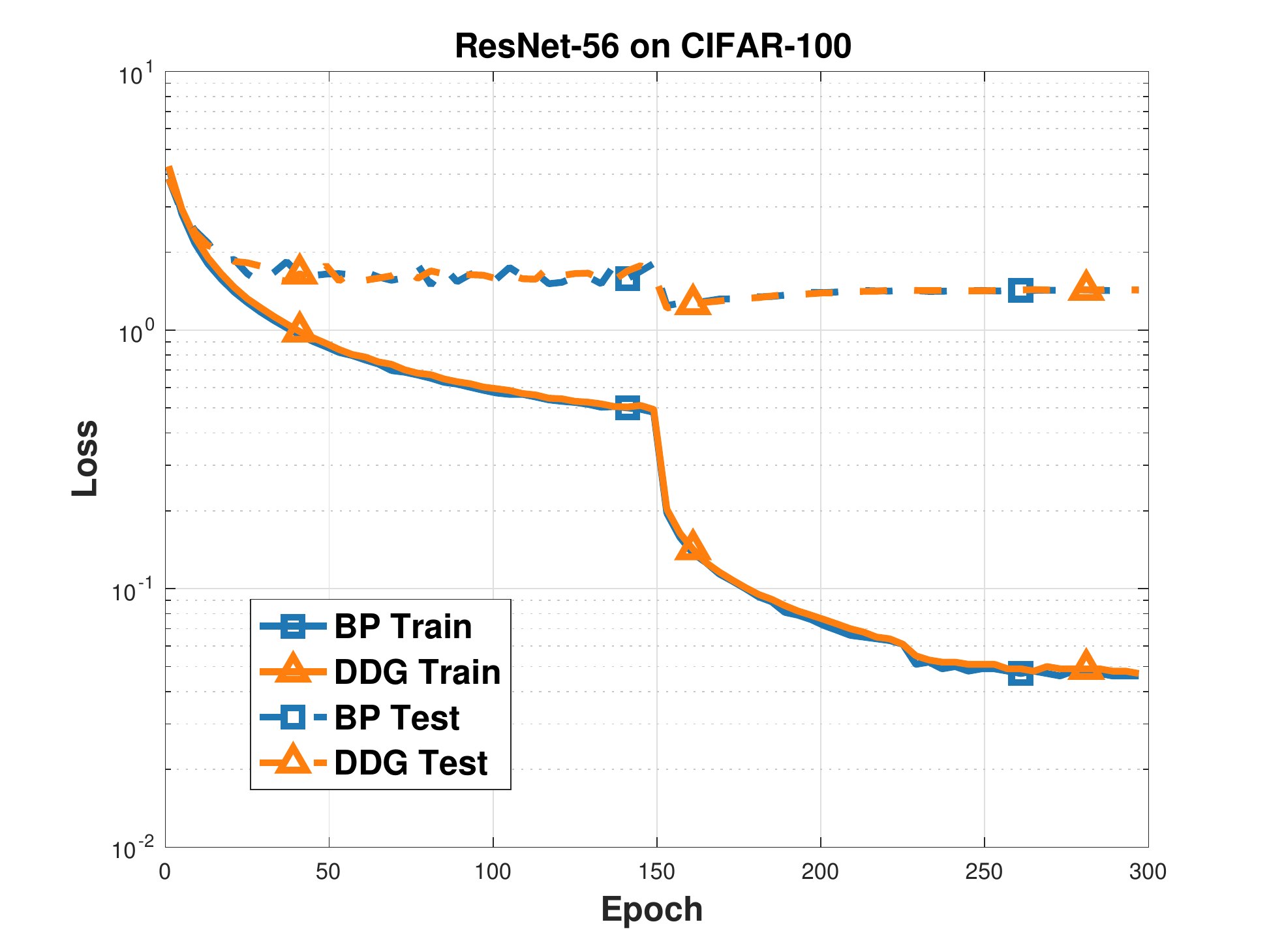}
	\end{subfigure}
	\begin{subfigure}[b]{0.24\textwidth}
		\centering
		\includegraphics[width=1.82in]{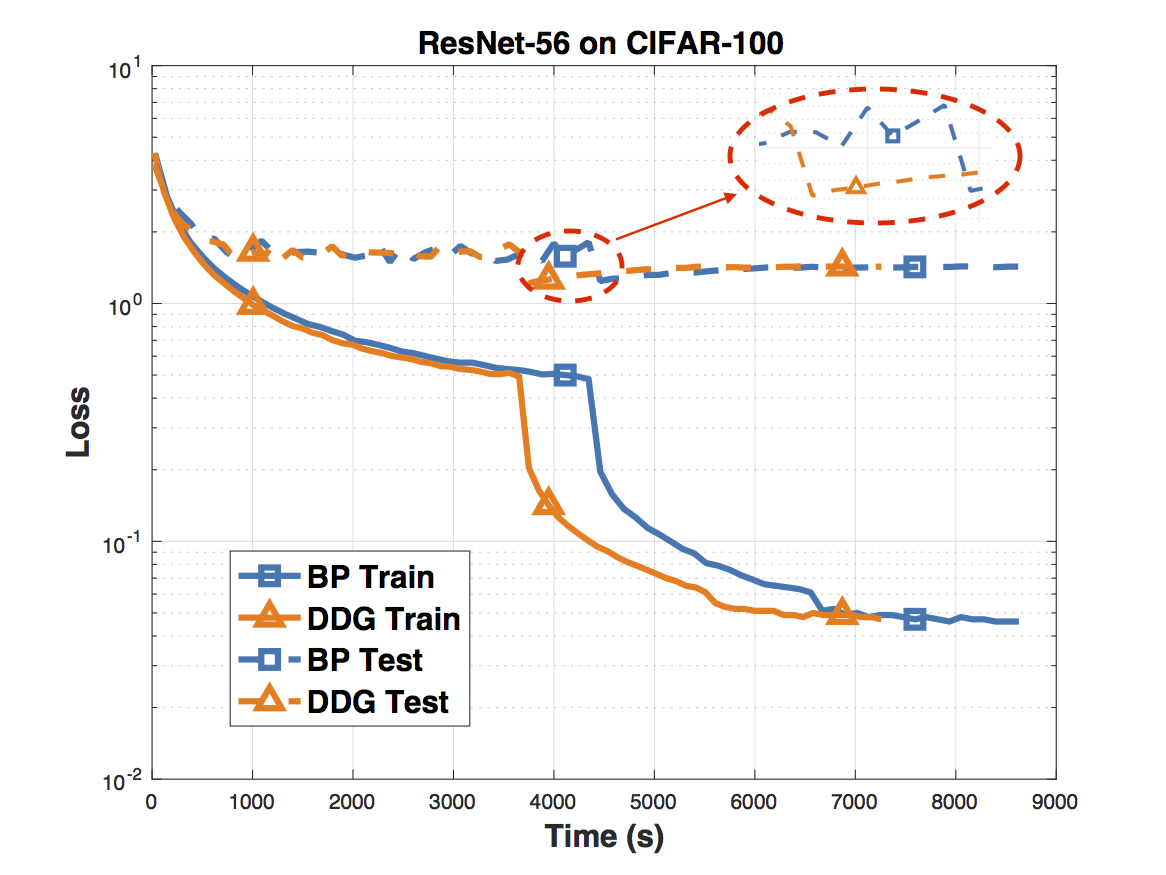}
	\end{subfigure}
	\begin{subfigure}[b]{0.24\textwidth}
		\centering
		\includegraphics[width=1.82in]{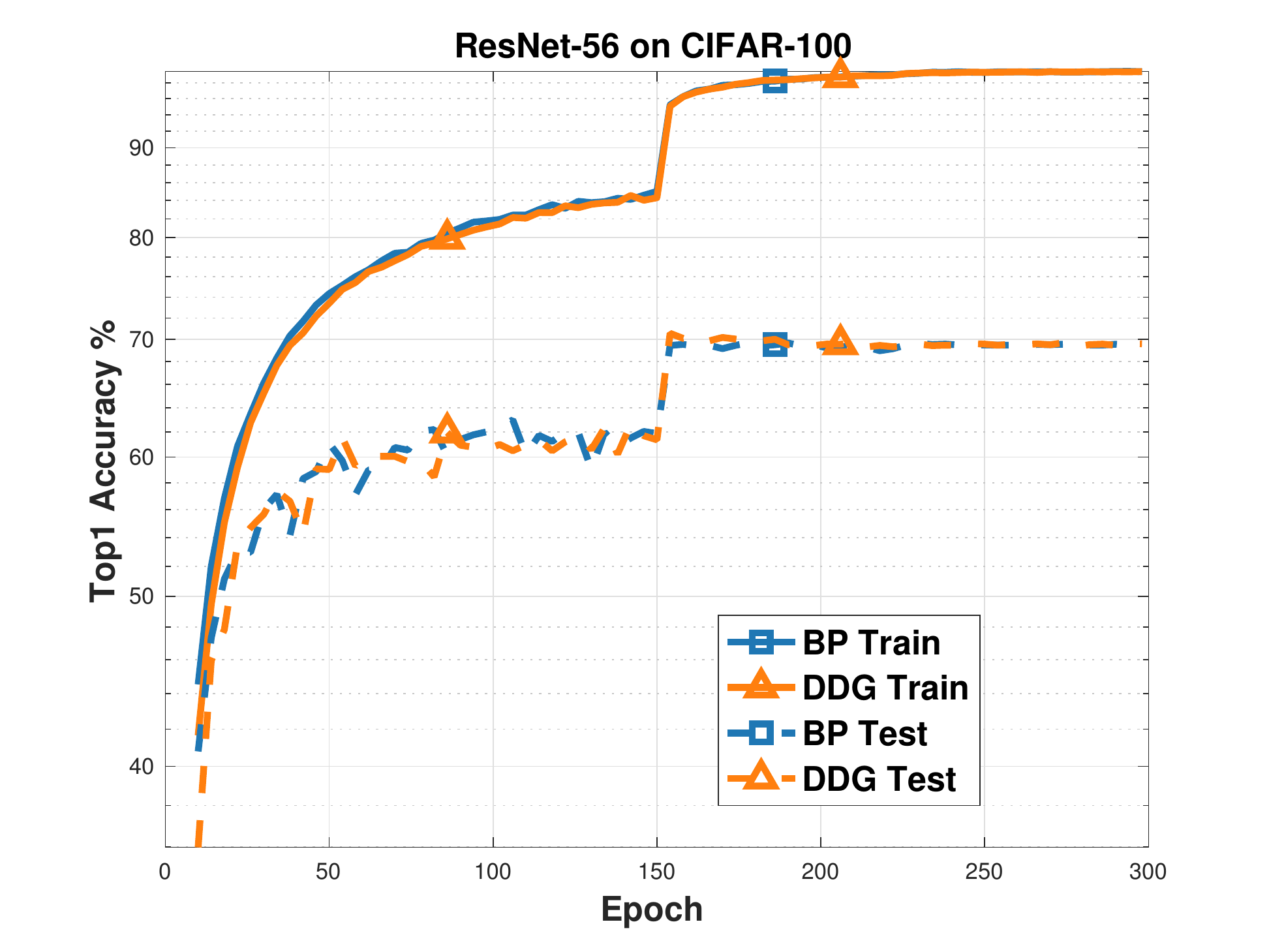}
	\end{subfigure}
	\begin{subfigure}[b]{0.24\textwidth}
		\centering
		\includegraphics[width=1.82in]{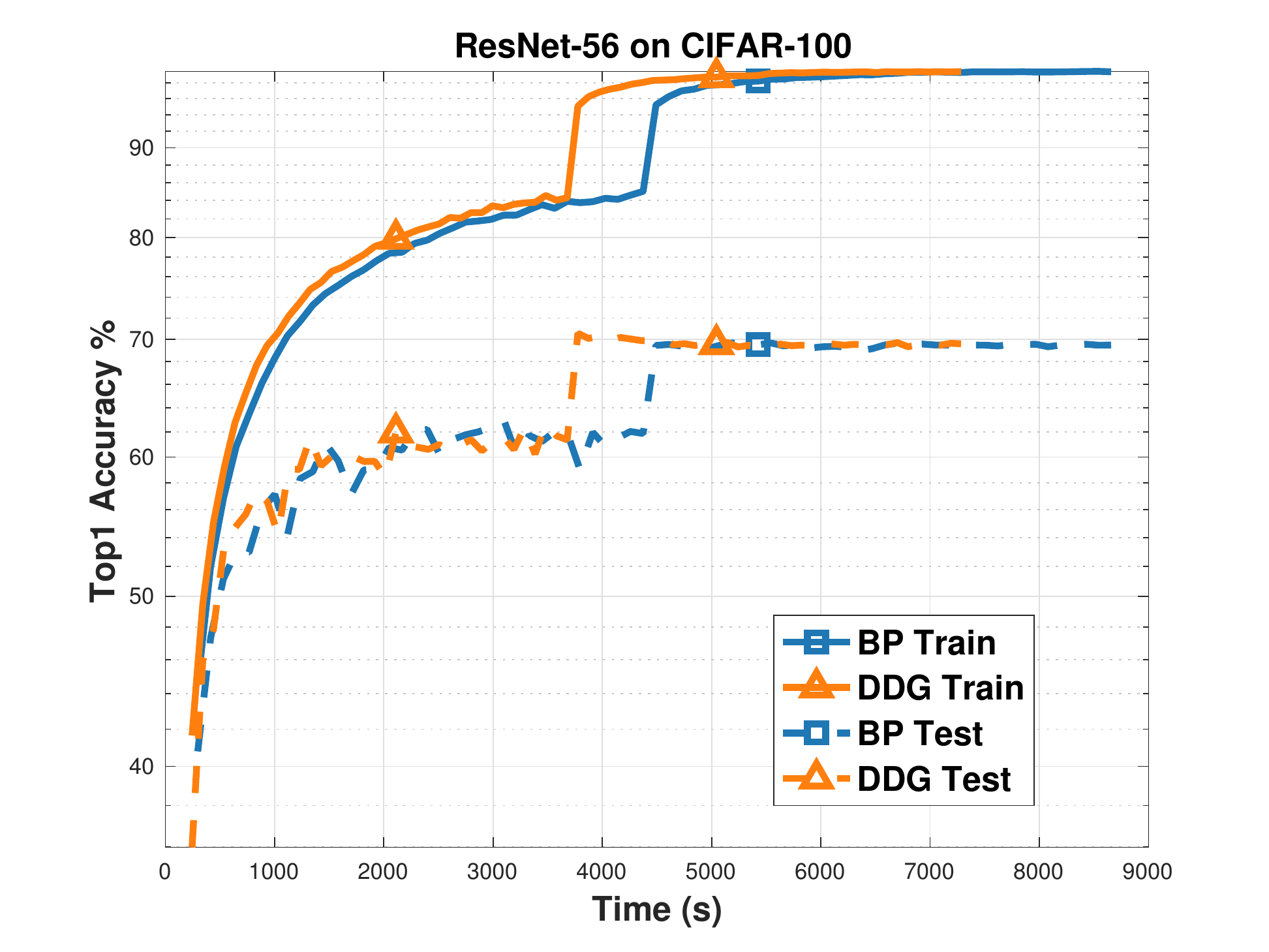}
	\end{subfigure}
	\begin{subfigure}[b]{0.24\textwidth}
		\centering
		\includegraphics[width=1.82in]{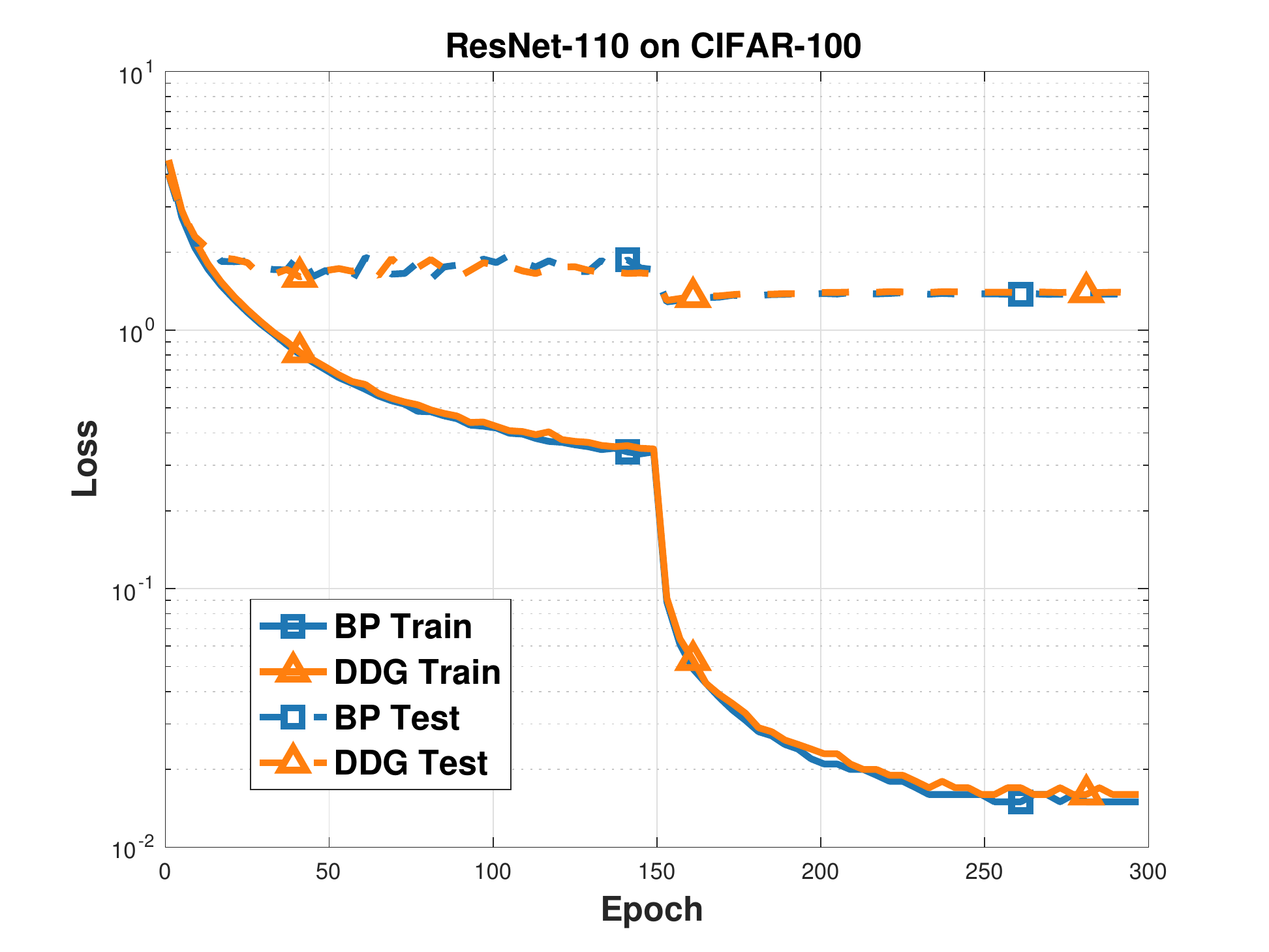}
	\end{subfigure}
	\begin{subfigure}[b]{0.24\textwidth}
		\centering
		\includegraphics[width=1.82in]{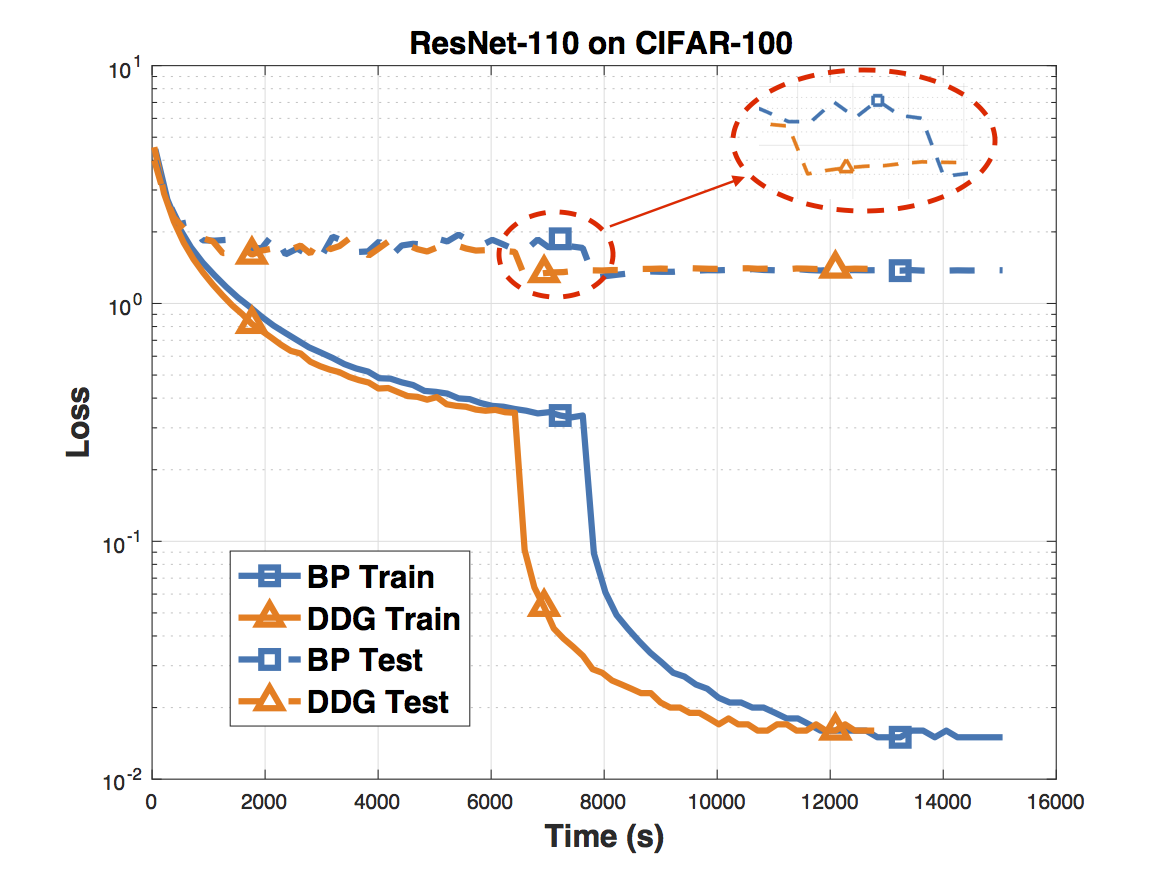}
	\end{subfigure}
	\begin{subfigure}[b]{0.24\textwidth}
		\centering
		\includegraphics[width=1.82in]{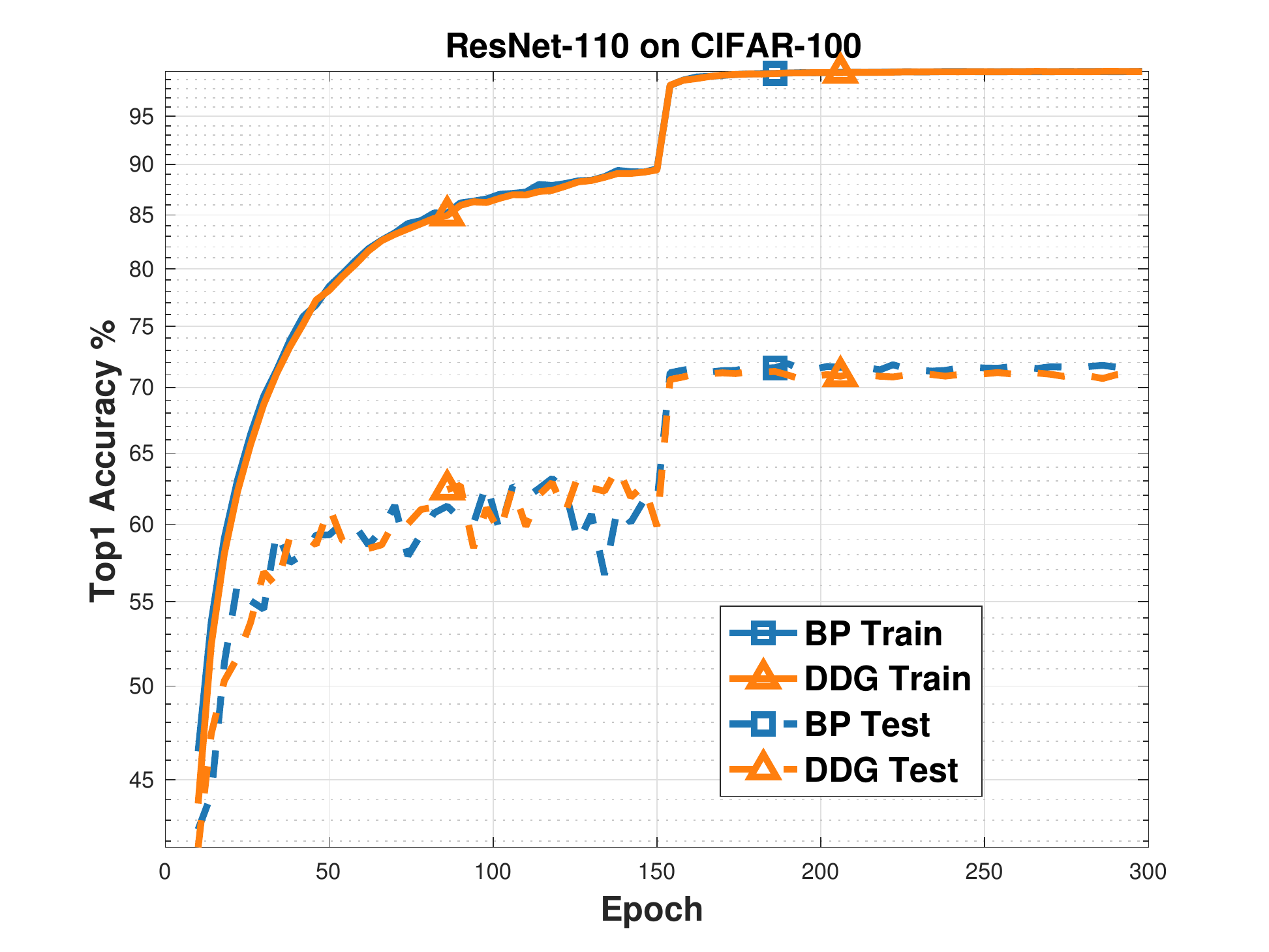}
	\end{subfigure}
	\begin{subfigure}[b]{0.24\textwidth}
		\centering
		\includegraphics[width=1.82in]{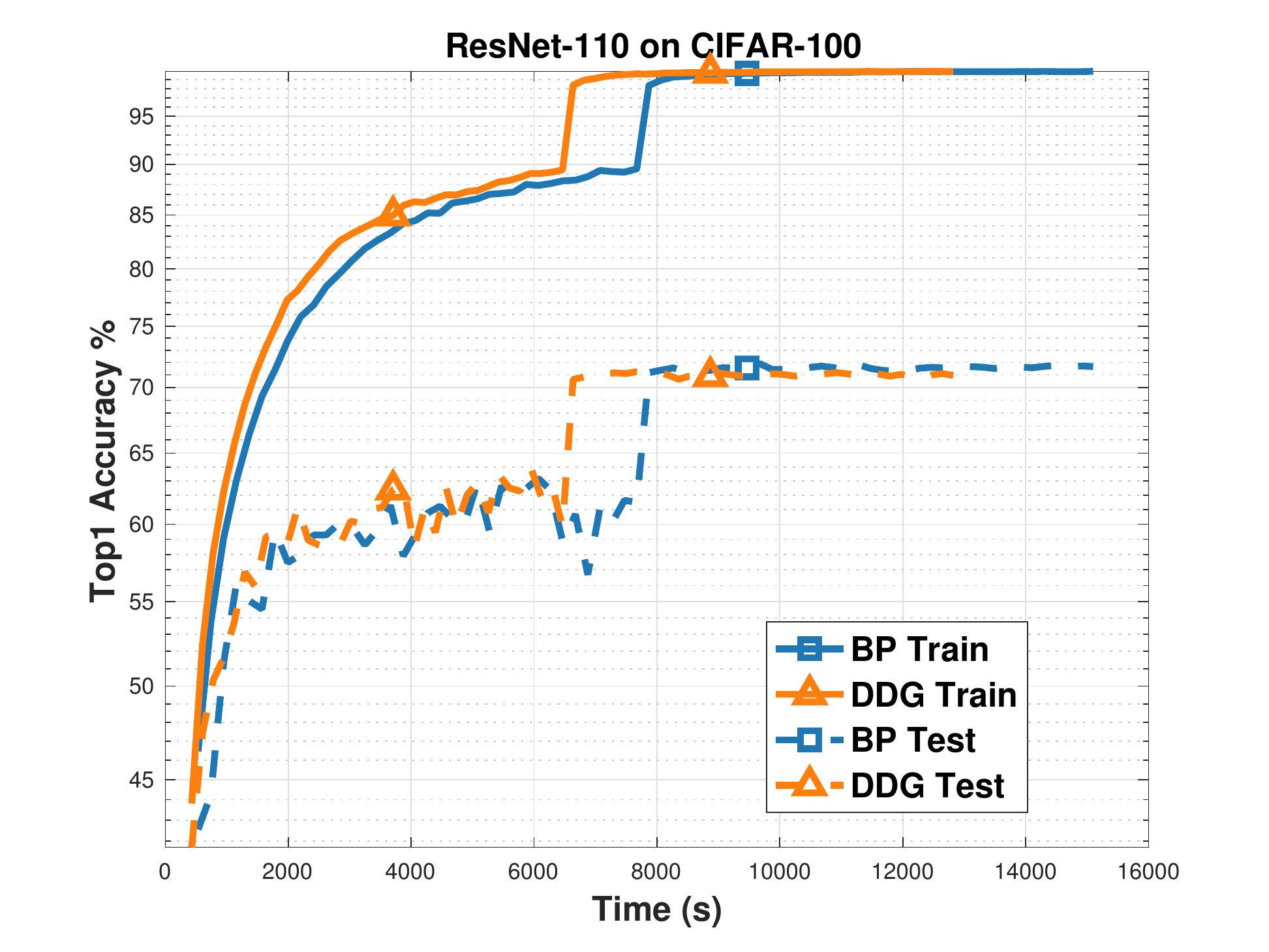}
	\end{subfigure}
	\caption{Training and testing curves for ResNet-56  and ResNet-110 on CIFAR-10 and CIFAR-100. \textbf{Column $1$ and $2$} present the loss function value regrading epochs and computation time respectively;  \textbf{Column $3$ and $4$}  present the Top 1 classification accuracy regrading epochs and computation time. For DDG, there is only one split point at the center of the network.}
	\label{cifar100}
\end{figure*}

\subsection{Optimizing  Deeper   Neural  Networks}
\label{exp_deep}
\begin{table}[t]
	\caption{The best Top 1 classification accuracy ($\%$) for ResNet-56 and ResNet-110 on the test data of CIFAR-10 and CIFAR-100. }
		\center
	\begin{tabular}{c|c|c|c|c}
		\hline
		\multirow{2}{*}{ Architecture} & \multicolumn{2}{c|}{CIFAR-10}	 &  \multicolumn{2}{c}{CIFAR-100} \\ \cline{2-5}
		&  BP&DDG &BP  &DDG \\ 
		\hline 
		ResNet-56 & \textbf{93.12 } & 93.11 &69.79 & \textbf{70.17}\\  \hline
		ResNet-110 & \textbf{93.53 } & 93.41 & \textbf{71.90} & 71.39  \\ \hline
	\end{tabular}
	\label{accuracy}
\end{table}
In this section, we employ DDG to optimize two very deep neural networks (ResNet-56 and ResNet-110) on CIFAR-10  and CIFAR-100. Each network is split into two modules at the center. We use SGD with the momentum of $0.9$ and the stepsize is initialized to $0.01$. Each model is trained for $300$ epochs and the stepsize is divided by a factor of $10$ at $150$ and $225$ epochs. The weight decay constant is set to $5 \times 10^{-4}$. We perform the same data augmentation as in section \ref{exp_classification}.  Experiments are run on a single Titan X GPU.

Figure \ref{cifar100} presents the experimental results of BP and DDG. We do not compare DNI because its performance is far worse when models are deep. Figures in the first column present the convergence of loss regarding epochs, showing that DDG and BP admit similar convergence rates. We can also observe that DDG converges faster when we compare the loss regarding computation time in the second column of Figure \ref{cifar100}.  In the experiment, the ``Volatile GPU Utility" is about $70\%$ when we train the models with BP.  Our method runs on two subprocesses such that it fully leverages the computing capacity of the GPU. 
 We can draw similar conclusions when we compare the Top 1 accuracy in the third and fourth columns of Figure \ref{cifar100}. 
 In Table \ref{accuracy}, we list the best Top 1 accuracy on the test data of CIFAR-10 and CIFAR-100. We can observe that DDG can obtain comparable or better  accuracy even when the network is deep.

\subsection{Scaling the Number of GPUs}

\label{distributed_training}
In this section, we split ResNet-110 into $K$ modules and allocate them across $K$ Titan X GPUs sequentially. We do not consider filter-wise model parallelism in this experiment. 
The selections of the parameters in the experiment are similar to Section \ref{exp_deep}. 
From Figure \ref{dist_conv}, we know that training networks in multiple GPUs does not affect the convergence rate. For comparison, we also count the computation time of backpropagation algorithm on a single GPU. The computation time is worse when we run backpropagation algorithm on multiple GPUs because of the communication overhead. In Figure \ref{dist2}, we can observe that forward time only accounts for about $32\%$ of the total computation time for backpropagation algorithm. Therefore,  backward locking is the main bottleneck. In Figure \ref{dist2}, it is obvious that when we increase the number of GPUs  from $2$ to $4$,  our method reduces about $30\%$ to $50\%$ of the total computation time. In other words, DDG achieves  a speedup of about $2$ times without loss of accuracy when we train the networks across $4$ GPUs.

\section{Conclusion}
In this paper, we propose decoupled parallel backpropagation algorithm, which breaks the backward locking in backpropagation algorithm using delayed gradients.  We then apply the decoupled parallel backpropagation to two stochastic methods for deep learning optimization. In the theoretical section, we also provide  convergence analysis and prove that  the proposed method guarantees convergence to critical points for the non-convex problem. Finally, we perform experiments on deep convolutional neural networks, results verifying that our method can accelerate the training significantly without loss of accuracy.

\section*{Acknowledgement}
This work was partially supported by U.S. NIH R01
AG049371, NSF IIS 1302675, IIS 1344152, DBI 1356628,
IIS 1619308, IIS 1633753.
\bibliography{example_paper}
\bibliographystyle{icml2018}

\appendix

\onecolumn
\icmltitle{Supplementary Materials for Paper ``Decoupled Parallel Backpropagation with Convergence Guarantee"}

\section{Proof to Lemma \ref{lem1}}
\begin{proof}
	Because the gradient of $f(w)$ is Lipschitz continuous in Assumption \ref{lips},  the following inequality holds that:
	\begin{eqnarray}
	f(w^{t+1}) &\leq & f(w^t) + \nabla f(w^t)^T \left( w^{t+1} - w^{t} \right) + \frac{L}{2} \left\| w^{t+1} - w^t\right\|^2_2.
	\end{eqnarray}
	From the update rule in Algorithm \ref{alg},  we take  expectation on both sides and obtain:
	\begin{eqnarray}
	\mathbb{E} \left[ f(w^{t+1}) \right] &\leq & f(w^t) - \gamma_t \mathbb{E} \left[ \nabla f(w^t)^T \left( \sum\limits_{k=1}^K \nabla f_{{\mathcal{G}(k)}, x_{i(t-K+k)}}\left( w^{t-K+k}\right)  \right)   \right] 
	\nonumber \\
	&&+ \frac{L\gamma_t^2}{2} \mathbb{E}\left\|\sum\limits_{k=1}^K \nabla f_{{\mathcal{G}(k)}, x_{i(t-K+k)}} (w^{t-K+k})\right\|^2_2 \nonumber \\
	&\leq & f(w^t) - \gamma_t \sum\limits_{k=1}^K \nabla f(w^t)^T \left( \nabla f_{\mathcal{G}(k)}\left( w^{t-K+k} \right)  + \nabla f_{\mathcal{G}(k)}\left(w^t\right) - \nabla f_{\mathcal{G}(k)}\left(w^t\right)  \right)  \nonumber \\
	&& +  \frac{L\gamma_t^2}{2}   \mathbb{E} \left\|\sum\limits_{k=1}^K   \nabla f_{{\mathcal{G}(k)}, x_{i(t-K+k)}} (w^{t-K+k}) - \nabla f(w^t) + \nabla f(w^t)  \right\|^2_2 \nonumber \\
	&= & f(w^t) - \gamma_t  \left\| \nabla f(w^t) \right\|^2_2  - \gamma_t \sum\limits_{k=1}^K \nabla f(w^t)^T \left( \nabla f_{{\mathcal{G}(k)}}\left( w^{t-K+k} \right) - \nabla f_{\mathcal{G}(k)}\left(w^t\right)  \right) \nonumber\\
	&&+ \frac{L\gamma_t^2}{2} \left\| \nabla f(w^t) \right\|^2_2 + \frac{L\gamma_t^2}{2} \mathbb{E} \left\| \sum\limits_{k=1}^K \nabla f_{{\mathcal{G}(k)}, x_{i(t-K+k)}} (w^{t-K+k}) - \nabla f(w^t) \right\|^2_2  \nonumber \\
	&& +  {L\gamma_t^2}  \sum\limits_{k=1}^K \nabla f(w^t)^T \left( \nabla f_{{\mathcal{G}(k)}}\left( w^{t-K+k} \right) - \nabla f_{\mathcal{G}(k)}\left(w^t\right)  \right) \nonumber \\
	&= & f(w^t) - \left(\gamma_t -\frac{L\gamma_t^2}{2}\right)  \left\| \nabla f(w^t) \right\|^2_2  + \underbrace{ \frac{L\gamma_t^2}{2} \mathbb{E} \left\| \sum\limits_{k=1}^K \nabla f_{{\mathcal{G}(k)}, x_{i(t-K+k)}} (w^{t-K+k}) - \nabla f(w^t) \right\|^2 }_{Q_1} \nonumber\\
	&& \underbrace{- (\gamma_t - L\gamma_t^2)\sum\limits_{k=1}^K \nabla f(w^t)^T \left( \nabla f_{{\mathcal{G}(k)}}\left( w^{t-K+k} \right) - \nabla f_{\mathcal{G}(k)}\left(w^t\right)  \right) }_{Q_2},
	\label{iq_1001}
	\end{eqnarray}
	where the second inequality follows from the unbiased gradient $\mathbb{E} \left[ \nabla f_{x_i} (w)  \right] = \nabla f(w) $. Because of  $\|x+y\|^2_2 \leq 2\|x\|^2_2 + 2\|y\|^2_2$ and $xy \leq \frac{1}{2}\|x\|^2_2+\frac{1}{2}\|y\|^2_2$, we have the upper bound of $Q_1$ and $Q_2$ as follows:
	\begin{eqnarray}
	Q_1 &=& \frac{L\gamma_t^2}{2} \mathbb{E}  \left\| \sum\limits_{k=1}^K \nabla f_{{\mathcal{G}(k)}, x_{i(t-K+k)}} (w^{t-K+k}) - \nabla f(w^t) - \sum\limits_{k=1}^K \nabla f_{\mathcal{G}(k)}(w^{t-K+k}) + \sum\limits_{k=1}^K \nabla f_{\mathcal{G}(k)}(w^{t-K+k}) \right\|^2_2 \nonumber \\
	&\leq & L\gamma_t^2 \underbrace{  \mathbb{E} \left\| \sum\limits_{k=1}^K \nabla f_{{\mathcal{G}(k)}, x_{i(t-K+k)}} (w^{t-K+k}) - \sum\limits_{k=1}^K \nabla f_{\mathcal{G}(k)}(w^{t-K+k}) \right\|^2_2 }_{Q_3}+L\gamma_t^2 \underbrace{ { \left\| \sum\limits_{k=1}^K \nabla f_{\mathcal{G}(k)}(w^{t-K+k}) - \nabla f(w^t) \right\|^2_2}}_{Q_4}.
	\end{eqnarray}
	\begin{eqnarray}
	Q_2 &=&- (\gamma_t - L\gamma_t^2)\sum\limits_{k=1}^K \nabla f(w^t)^T \left( \nabla f_{{\mathcal{G}(k)}}\left( w^{t-K+k} \right) - \nabla f_{\mathcal{G}(k)}\left(w^t\right) \right) \nonumber\\
	&\leq & \frac{\gamma_t - L\gamma_t^2}{2} \left\| \nabla f(w^t) \right\|^2_2 + \frac{\gamma_t - L\gamma_t^2}{2} \left\| \sum\limits_{k=1}^K \nabla f_{\mathcal{G}(k)}(w^{t-K+k}) - \nabla f(w^t) \right\|^2_2.
	\end{eqnarray}
	As per the equation regarding variance $\mathbb{E}\|\xi - \mathbb{E}[\xi] \|^2_2 = \mathbb{E} \|\xi\|^2_2 - \|\mathbb{E}[\xi]\|^2_2$, we can bound $Q_3$ as follows:
	\begin{eqnarray}
	Q_3 &= & \mathbb{E} \left\| \sum\limits_{k=1}^K \nabla f_{{\mathcal{G}(k)}, x_{i(t-K+k)}} (w^{t-K+k}) - \sum\limits_{k=1}^K \nabla f_{\mathcal{G}(k)}(w^{t-K+k}) \right\|^2_2 \nonumber \\
	&= &  \sum\limits_{k=1}^K  \mathbb{E} \left\| \nabla f_{{\mathcal{G}(k)}, x_{i(t-K+k)}} (w^{t-K+k}) -  \nabla f_{\mathcal{G}(k)}(w^{t-K+k}) \right\|^2_2 \nonumber \\
	&\leq &  \sum\limits_{k=1}^K  \mathbb{E} \left\| \nabla f_{{\mathcal{G}(k)}, x_{i(t-K+k)}} (w^{t-K+k})  \right\|^2_2 \nonumber \\
	&\leq & KM,
	\end{eqnarray}
	where the equality follows from the definition of $\nabla f_{\mathcal{G}(k)}(w)$ such that $[\nabla f_{\mathcal{G}(k)}(w)]_j =0, \hspace{0.1cm} \forall j \notin {\mathcal{G}(k)}$ and the last inequality is from Assumption \ref{bg}. We can also get the upper bound of $Q_4$:
	\begin{eqnarray}
	Q_4 &=& \left\| \sum\limits_{k=1}^K \nabla f_{\mathcal{G}(k)}(w^{t-K+k}) - \nabla f(w^t) \right\|^2_2 \nonumber \\
	&= & \sum\limits_{k=1}^K  \left\|\nabla f_{\mathcal{G}(k)}(w^{t-K+k}) - \nabla f_{\mathcal{G}(k)}(w^t) \right\|^2_2 \nonumber \\
	&\leq & \sum\limits_{k=1}^K  \left\|\nabla f(w^{t-K+k}) - \nabla f(w^t) \right\|^2_2 \nonumber \\
	&\leq & L^2\sum\limits_{k=1}^K  \left\| \sum\limits_{j=\max\{0, t-K+k\}}^{t-1} \left(w^{j+1} -  w^j \right) \right\|^2 \nonumber \\ 
	&\leq & L^2 \gamma_{\max\{0, t-K+1\}}^2 K\sum\limits_{k=1}^K  \sum\limits_{j=\max\{0, t-K+k\}}^{t-1}  \left\|  \sum\limits_{k=1}^K \nabla f_{{\mathcal{G}(k)}, x_{(j)}} \left(w^{j-K+k}\right)   \right\|^2_2 \nonumber \\ 
	&\leq & KL\gamma_t \frac{\gamma_{\max\{0, t-K+1\}}}{\gamma_t}\sum\limits_{k=1}^K  \sum\limits_{j=\max \{0,t-K+k\}}^{t-1}  \left\|  \sum\limits_{k=1}^K \nabla f_{{\mathcal{G}(k)}, x_{(j)}} \left(w^{j-K+k}\right)   \right\|^2_2 \nonumber \\ 
	&\leq & L\gamma_t \sigma K^4  M,
	\end{eqnarray}
	where the second inequality is from Assumption \ref{lips},  the fourth inequality follows from that $L\gamma_t \leq 1$ and  the last inequality follows from $	\| z_1 + ... + z_r\|^2_2   \leq  r (\|z_1\|^2_2 +...+ \|z_r\|^2_2 )$, Assumption \ref{bg} and $\sigma:= \max_t\frac{\gamma_{\max\{0, t-K+1\}}}{\gamma_t}$.
	Integrating the upper bound of $Q_1$, $Q_2$, $Q_3$ and $Q_4$ in  (\ref{iq_1001}),  we have:
	\begin{eqnarray}
	\mathbb{E} \left[ f(w^{t+1}) \right]  - f(w^t)&\leq & -\frac{\gamma_t}{2}\left\| \nabla f(w^t) \right\|^2_2  + \gamma_t^2 L  \sum\limits_{k=1}^K  \mathbb{E} \left\| \nabla f_{{\mathcal{G}(k)}, x_{i(t-K+k)}} (w^{t-K+k})  \right\|^2 \nonumber\\
	&& + \frac{\gamma_t + L\gamma_t^2}{2} \left\| \sum\limits_{k=1}^K \nabla f_{\mathcal{G}(k)}(w^{t-K+k}) - \nabla f(w^t) \right\|^2_2. \nonumber \\
	&\leq& -\frac{\gamma_t}{2}  \left\| \nabla f(w^t) \right\|^2  +  \gamma_t^2 L M_K,
	\end{eqnarray}
	where we let $M_K =KM +    \sigma K^4 M$. 
\end{proof}

\section{Proof to Theorem \ref{them1}}
\begin{proof}
	When $\gamma_t$ is constant and $\gamma_t = \gamma$, taking total expectation of (\ref{lem1_iq2}) in Lemma \ref{lem1}, we obtain:
	\begin{eqnarray}
	\mathbb{E} \left[ f(w^{t+1}) \right]  - \mathbb{E}\left[ f(w^t) \right] \leq -\frac{\gamma}{2}  \mathbb{E} \left\| \nabla f(w^t) \right\|^2_2  +  \gamma^2 L M_K,
	\label{thm2_iq1}
	\end{eqnarray}
	where $\sigma = 1$ and $M_K =KM +    K^4 M$.
	Summing (\ref{thm2_iq1}) from $t=0$ to $T-1$, we have:
	\begin{eqnarray}
	\mathbb{E} \left[ f(w^{T}) \right]  - f(w^0)&\leq & -\frac{\gamma}{2} \sum\limits_{t=0}^{T-1}  \mathbb{E}\left\| \nabla f(w^t) \right\|^2_2  + T\gamma^2 L M_K.
	\end{eqnarray}
	Suppose $w^*$ is the optimal solution for $f(w)$, therefore $f(w^*) - f(w^0) \leq \mathbb{E} \left[f(w^T) \right]- f(w^0) $. Above all, the following inequality is guaranteed that:
	\begin{eqnarray}
	\frac{1}{T} \sum\limits_{t=0}^{T-1}\mathbb{E}  \left\| \nabla f(w^t) \right\|^2_2 \leq \frac{2\left( f(w^0) - f(w^*) \right)}{\gamma T} + {2\gamma L M_K}.
	\end{eqnarray} 
	
\end{proof}

\section{Proof to Theorem \ref{them2}}
\begin{proof} 
	$\{\gamma_t\} $ is a diminishing sequence and $\gamma_t = \frac{\gamma_0}{1+t}$, such that $\sigma \leq K$ and $M_K =KM +    K^5 M$. Taking total expectation of (\ref{lem1_iq2}) in Lemma \ref{lem1} and summing it from $t=0$ to $T-1$, we obtain:
	\begin{eqnarray}
	\mathbb{E} \left[ f(w^{T}) \right]  - f(w^0)&\leq & -\frac{1}{2} \sum\limits_{t=0}^{T-1} \gamma_t \mathbb{E}\left\| \nabla f(w^t) \right\|^2_2  + \sum\limits_{t=0}^{T-1} \gamma_t^2 L M_K.
	\end{eqnarray}
	Suppose $w^*$ is the optimal solution for $f(w)$, therefore $f(w^*) - f(w^0) \leq \mathbb{E} \left[f(w^T) \right]- f(w^0) $. Letting $\Gamma_T = \sum\limits_{t=0}^{T-1} \gamma_t$, we have: 
	\begin{eqnarray}
	\frac{1}{\Gamma_T} \sum\limits_{t=0}^{T-1} \gamma_t\mathbb{E} \left\| \nabla f(w^t) \right\|^2_2 \leq \frac{2\left( f(w^0) - f(w^*) \right)}{\Gamma_T} + \frac{2\sum\limits_{t=0}^{T-1} \gamma_t^2 L M_K}{\Gamma_T}.
	\end{eqnarray} 
	We complete the proof.
	
\end{proof}

\end{document}